\documentclass[11pt, a4paper, onecolumn]{paistechreport}

\usepackage{algorithm}
\usepackage{algorithmic}
\usepackage{multirow}
\usepackage[authoryear, sort&compress, round]{natbib}
\usepackage{silence}
\renewcommand{\today}{12 March 2026}

\title{Self-Flow-Matching assisted Full Waveform Inversion}
\correspondingauthor{huang26@seas.upenn.edu}

\author[1]{Xinquan~Huang}
\author[1]{Paris~Perdikaris}
\affil[1]{University of Pennsylvania}

\begin{abstract}
Full-waveform inversion (FWI) is a high-resolution seismic imaging method that estimates subsurface velocity by matching simulated and recorded waveforms. However, FWI is highly nonlinear, prone to cycle skipping, and sensitive to noise, particularly when low frequencies are missing or the initial model is poor, leading to failures under imperfect acquisition. Diffusion-regularized FWI introduces generative priors to encourage geologically realistic models, but these priors typically require costly offline pretraining and can deteriorate under distribution shift.
Moreover, they assume Gaussian initialization and a fixed noise schedule, in which it is unclear how to map a deterministic FWI iterate and its starting model to a well-defined diffusion time or noise level. 
To address these limitations, we introduce Self-Flow-Matching assisted Full-Waveform Inversion (SFM-FWI), a physics-driven framework that eliminates the need for large-scale offline pretraining while avoiding the noise-level alignment ambiguity.
SFM-FWI leverages flow matching to learn a transport field without assuming Gaussian initialization or a predefined noise schedule, so the initial model can be used directly as the starting point of the dynamics.
Our approach trains a single flow network online using the governing physics and observed data. At each outer iteration, we build an interpolated model and update the flow by backpropagating the FWI data misfit, providing self-supervision without external training pairs. 
Experiments on challenging synthetic benchmarks show that SFM-FWI delivers more accurate reconstructions, greater noise robustness, and more stable convergence than standard FWI and pretraining-free regularization methods.
\end{abstract}

\begin{document}
\maketitle

\section{Introduction}
Full-waveform inversion (FWI) \citep{Tarantola1984} is widely recognized as a high-resolution seismic inversion technique by minimizing the misfit between observed and simulated seismic data, and has been extensively applied in oil and gas exploration
\citep{Pratt1990,Virieux2009} and global seismology \citep{Dessa2004,tromp2020seismic}. 
However, due to its severe nonlinearity and non-convexity \citep{Virieux2009}, it suffers from cycle skipping \citep{Bunks1995,sirgue2004efficient,symes2008migration} and is highly sensitive to noise and observation errors \citep{Brossier2010,Warner2013}.
Those imperfect observations and a poor starting model often limited FWI's ability to achieve high-resolution inversion results.
To mitigate these issues, various strategies have been proposed, including multi-scale frequency continuation \citep{Bunks1995,fichtner2013multiscale}, robust data-misfit functions like optimal transport objective function \citep{Metivier2016}, or the weighted
envelope correlation misfit function \citep{song2023weighted}, model regularization \citep{xiang2016efficient}, and incorporating prior model information into the inversion \citep{wang_prior_2023}.
These strategies are largely complementary.
Here, we focus on regularization and prior-based formulations, since they provide a principled way to inject geological plausibility into PDE-constrained optimization and can be combined with continuation or robust misfit objectives when needed.
Many handcrafted regularization methods, including Tikhonov \citep{Aster2011} and total variation (TV) regularizations \citep{alkhalifah2018full,xiang2016efficient}, often lead to models that are either smooth or piecewise smooth, respectively.
Sparsity-promoting formulations using curvelet, seislet, or learned dictionaries have also been investigated to stabilize inversion \citep{Aster2011,Lix2012,Xue2017,huang2019robust}. 
Although effective, these methods rely on manually designed priors and hyperparameter tuning, and may oversimplify complex geological structures.

With the success of deep learning across applications such as pattern recognition, control, language modeling, and image generation \citep{Lecun2015}, machine learning techniques have been increasingly incorporated into FWI. 
Based on how they are incorporated, existing learning-assisted regularization can be broadly categorized into explicit and implicit regularization strategies. 
In explicit regularization, additional prior information is incorporated directly into the physics-based objective, for instance, by augmenting the data-misfit with a handcrafted or learned regularizer or by enforcing an explicit constraint set where regularization encodes geological plausibility (e.g., learned plug-and-play denoisers or generative priors such as GANs and diffusion models) \citep{wang_prior_2023,sun2023full,mosser2020stochastic,venkatakrishnan2013plug}. 
Implicit regularization keeps the data-misfit formulation unchanged but biases the solution through deep neural reparameterization, such as implicit neural representations or deep image prior-style reparameterization, which can suppress noise-like components and promote coherent structures without explicitly introducing a penalty term \citep{ulyanov2018deep,wu2019parametric,zhu2022integrating,sun2023implicit,wu2025does}. 
These developments collectively aim to enhance the robustness of FWI under practical acquisition and modeling limitations, while retaining consistency with wave physics.

Recently, inspired by advances in generative modeling, diffusion-based priors have been introduced into FWI \citep{wang_prior_2023,wang_geological_2025}. 
By constraining inversion to evolve within a learned distribution of realistic velocity models, diffusion-regularized FWI has shown improved stability and artifact suppression compared to classical regularization in a range of settings.
However, diffusion models introduce an intrinsic conceptual and practical mismatch when coupled with PDE-constrained inversion. 
Standard diffusion generative models are formulated as reverse stochastic processes that start from a Gaussian distribution and progressively denoise according to a predefined noise schedule \citep{ho_denoising_2020,song_score-based_2021}. 
When integrated into FWI, this requires interpreting the current velocity model as a noisy sample at a specific diffusion time step. 
In practice, however, the initial model in FWI is neither a Gaussian realization nor associated with a well-defined noise level. 
This ambiguity in aligning inversion iterations with diffusion time leads to inconsistencies in initialization and potential instability. 
Moreover, diffusion models typically require large-scale offline training on representative velocity datasets, and performance may degrade under distribution shift between training data and the target geological setting \citep{wang_geological_2025}.
Obtaining massive velocity models is challenging.
\cite{luo_self-diffusion_2025} proposed a self-diffusion approach for the linear inverse problem, which can avoid heavy pretraining of the diffusion models, but is not suitable for this non-linear inverse problem.

To alleviate these limitations, we propose \emph{\textbf{S}elf-\textbf{F}low-\textbf{M}atching assisted \textbf{F}ull \textbf{W}aveform \textbf{I}nversion} (\textbf{SFM-FWI}), a deterministic generative-prior framework based on flow matching. 
Unlike diffusion models, flow matching does not impose a predefined assumption on the source distribution. 
Instead, it learns a continuous transport field that maps samples from an arbitrary source distribution to a target distribution through an ordinary differential equation (ODE). 
Hence, the current inversion model can naturally be regarded as the source distribution, without requiring artificial Gaussian corruption or predefined noise schedules, which is particularly well suited for FWI. 
By coupling short physics-based FWI refinements with online flow matching, SFM-FWI learns a velocity-model transport field directly driven by observed data and governing wave physics.

The resulting inversion trajectory follows a stable coarse-to-fine evolution, automatically prioritizing low-wavenumber reconstruction before introducing high-wavenumber details. 
Importantly, the proposed method eliminates the need for massive pretrained generative datasets and avoids stochastic reverse sampling, while retaining the benefits of manifold-constrained model evolution. 
Extensive experiments on synthetic benchmarks demonstrate that SFM-FWI achieves higher accuracy, improved robustness to noise, and more stable convergence than classical FWI and other pretraining-free regularized approaches.

Our contributions are summarized as follows:
\begin{itemize}
    \item We introduce a flow-matching-regularized FWI framework that eliminates the need for Gaussian initialization and predefined diffusion schedules, enabling direct transport from the current inversion model.
    \item We design a self-supervised online training scheme that guides flow matching with physics-based FWI updates, requiring no paired velocity datasets for pretraining.
    \item We demonstrate through comprehensive experiments that SFM-FWI achieves superior reconstruction accuracy, robustness, and convergence stability compared to classical and pretraining-free baselines.
\end{itemize}

The rest of this paper is organized as follows. We first introduce the basic formulations of FWI and then describe our methods in Section \ref{sec:method}. 
To demonstrate the effectiveness of the proposed method, Section \ref{sec:experiments} presents the experiments on various challenging velocity models. 
Finally, we discuss in Section \ref{sec:discussion} and conclude in Section \ref{sec:conclusion}.

\section{Methodology}
\label{sec:method}

In this section, we first present the standard time-domain acoustic FWI formulation and introduce flow matching.
Then, we develop SFM-FWI, which couples a self-supervised-trained flow model with physics-based inversion to improve the stability and robustness of the inversion.
\begin{figure*}[!htb]
    \centering
    \includegraphics[width=0.8\linewidth]{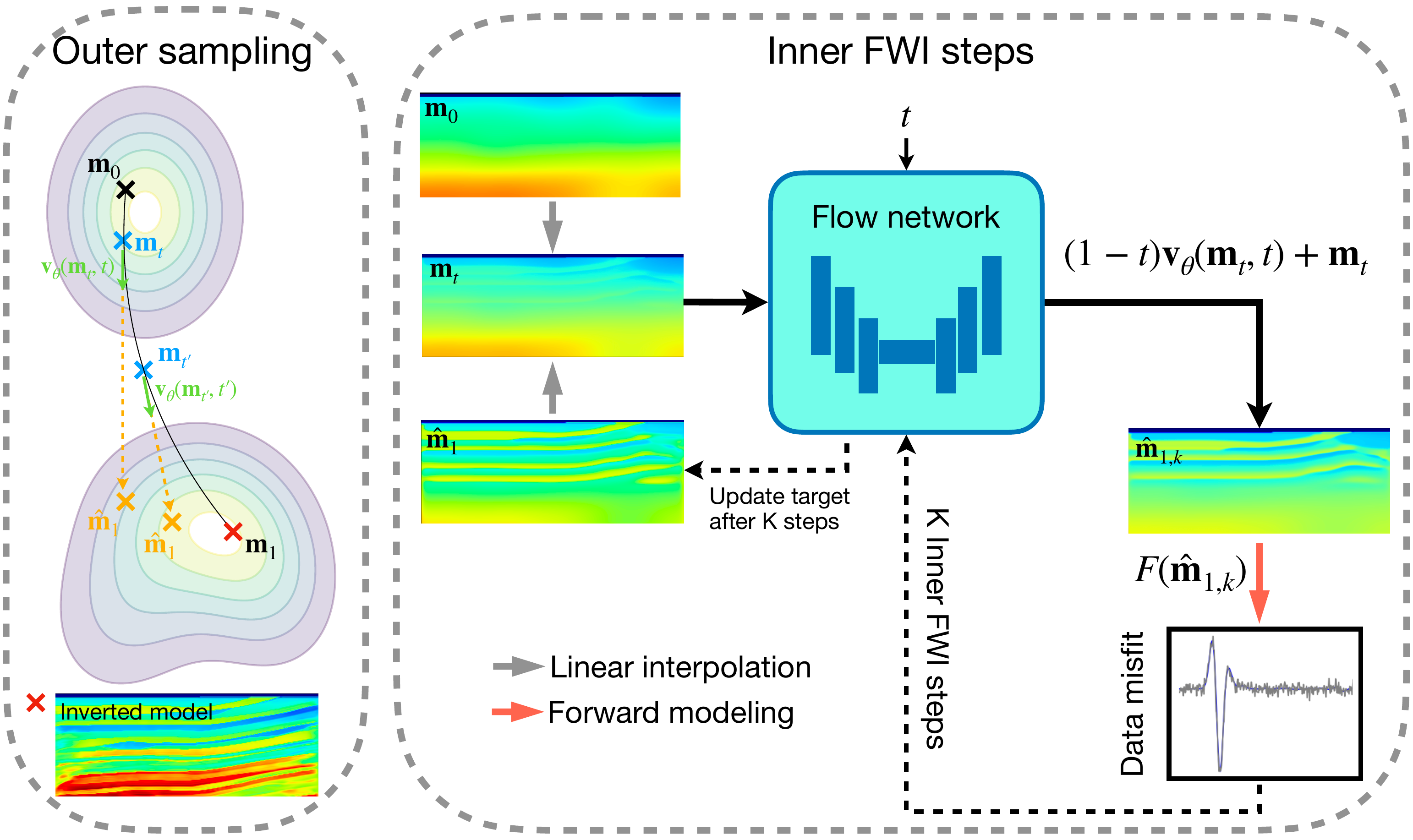}
    \caption{SFM-FWI schematic. The method alternates between outer sampling and inner misfit-driven online training of a flow network to update \(\hat{\mathbf{m}}_1\). This produces a stable coarse-to-fine inversion without pretrained priors.}
    \label{fig:diagram}
\end{figure*}

\subsection{Full waveform inversion}
\label{sec:fwi}

FWI aims to estimate subsurface model parameters by matching simulated seismic seismograms to observed ones.
Let \(\mathbf{m}(\mathbf{x})\) denote the model parameters (seismic velocity in this paper), \(\mathbf{d}_{\mathrm{obs}}\) the recorded data, and \(F(\cdot)\) the forward modeling operator that generates synthetic data \(\mathbf{d}_{\mathrm{syn}} = F(\mathbf{m})\).
FWI is commonly formulated as a wave-equation-constrained optimization problem \citep{Tarantola1984,Virieux2009,fichtner2010full}:
\begin{equation}
\min_{\mathbf{m}}\ \Phi(\mathbf{m})
:= \frac{1}{2}\,\|F(\mathbf{m})-\mathbf{d}_{\mathrm{obs}}\|_2^2 \ +\ \lambda R(\mathbf{m}),
\label{eq:fwi_obj}
\end{equation}
where the first term is the data-misfit objective, and \(R(\mathbf{m})\) is an optional regularization term encoding prior information with weight \(\lambda\).

Regarding \(F(\cdot)\), taking the acoustic wave equation as an example, the variable-density acoustic wave equation for the pressure field \(u(x,z,t)\) is formulated as
\begin{equation}
\begin{aligned}
\frac{1}{\rho(x,z)\,\mathbf{m}^2(x,z)}\,\frac{\partial^2u(x,z,t)}{\partial t^2}
\;-\;\nabla\!\left(\frac{1}{\rho(x,z)}\nabla u(x,z,t)\right)
\;\\=\; f(x,z,t)\delta(x-x_s,z-z_s),
\label{eq:vd_acoustic}
\end{aligned}
\end{equation}
where \(\rho(x,z)\) denotes the density, \((x_s,z_s)\) is the source location, \(f\) is the source wavelet, and \(\mathbf{m}(x,z)\) is the velocity value of \(\mathbf{m}\) at \((x,z)\).
In this study, we assume constant density \(\rho(x,z)\equiv\rho_0\).
Synthetic data \(\mathbf{d}_{\mathrm{syn}}\) are obtained by sampling the wavefield at receiver locations.

To minimize \(\Phi(\mathbf{m})\), we apply the adjoint-state method \citep{Plessix2006} to compute gradients efficiently.
Let the data residual be \(\mathbf{r} = \mathbf{d}_{\mathrm{syn}}-\mathbf{d}_{\mathrm{obs}}\).
An adjoint wavefield is generated by back-propagating the residual, and the gradient is formed by a space--time correlation of forward $u(\cdot,t)$ and adjoint wavefields $v(\cdot,t)$:
\begin{equation}
\nabla_{\mathbf{m}} \Phi(\mathbf{m})
\ \propto\ \int_0^T \mathcal{G}\big(u(\cdot,t),v(\cdot,t);\mathbf{m}\big)\,dt
\ +\ \lambda \nabla_{\mathbf{m}} R(\mathbf{m}),
\label{eq:fwi_grad}
\end{equation}
where \(\mathcal{G}(\cdot)\) denotes the parameterization-dependent imaging condition.
The model is updated by a gradient-based method
\begin{equation}
\mathbf{m}^{k+1} = \mathbf{m}^{k} - \alpha_k \nabla_{\mathbf{m}}\Phi(\mathbf{m}^k),
\label{eq:fwi_update}
\end{equation}
with step size \(\alpha_k\) determined by line search or heuristics.

\subsection{Flow matching}
\label{sec:flowmatching}

The nonconvexity of \(\Phi\) leads to cycle skipping, particularly when low frequencies are missing, illumination is limited, or starting models are poor, and can be further aggravated by noisy or imperfect data.
Recently, generative priors have been introduced to constrain inversion within a manifold of geologically plausible velocity models \citep{wang_prior_2023}.
However, diffusion-based priors are typically formulated as reverse processes that assume Gaussian initialization and rely on a predefined noise schedule.
When coupled with FWI, this introduces an intrinsic ambiguity because an FWI iterate is a data-conditioned estimate, and the initial model is neither a Gaussian sample nor naturally associated with a well-defined diffusion time (or noise level).
As a result, it is unclear how to consistently align the inversion state with a diffusion step during sampling.
This motivates the use of flow matching.
Unlike diffusion models, flow matching does not require a specific assumption on the source distribution.
Therefore, the current inversion model can be directly treated as the starting point of the transport dynamics, without artificial Gaussian corruption or noise-level calibration. 

Flow matching learns a continuous-time transport field \citep{lipman2022flow} that maps states from a source distribution to a target distribution by integrating an ODE.
Let \(\mathbf{x}(t)\in\mathbb{R}^{H\times W}\) be a time-dependent state and \(\mathbf{v}_\theta(\mathbf{x},t)\) be a neural vector field. The transport dynamics are
\begin{equation}
\frac{d \mathbf{x}(t)}{dt} = \mathbf{v}_\theta(\mathbf{x}(t),t), \qquad t\in[0,1].
\label{eq:fm_ode}
\end{equation}
A common formulation constructs endpoint pairs \((\mathbf{x}_0,\mathbf{x}_1)\) and samples intermediate states along a simple path
\begin{equation}
\mathbf{x}_t = (1-t)\mathbf{x}_0 + t \mathbf{x}_1, \qquad t\sim \mathcal{U}[0,1],
\label{eq:fm_path}
\end{equation}
whose path velocity is \(\frac{d\mathbf{x}_t}{dt} = \mathbf{x}_1-\mathbf{x}_0\).
Flow matching fits \(\mathbf{v}_\theta\) by minimizing
\begin{equation}
\mathcal{L}_{\mathrm{FM}}(\theta)
= \mathbb{E}_{t,\,\mathbf{x}_0,\,\mathbf{x}_1}\left[\left\|\mathbf{v}_\theta(\mathbf{x}_t,t) - (\mathbf{x}_1-\mathbf{x}_0)\right\|_2^2\right].
\label{eq:fm_loss}
\end{equation}

\subsection{Self-flow-matching assisted full waveform inversion}
\label{sec:sfmfwi}
As highlighted in prior studies \citep{wang_prior_2023,wang_geological_2025}, generative priors can enhance the robustness and accuracy of FWI.
A practical limitation, however, is that many high-capacity generative models require large collections of representative velocity models for offline training and careful training to generalize across geological regimes.
In seismic inversion, such datasets are expensive to curate, and a distribution shift between training data and the target survey area can noticeably degrade performance \citep{wang_geological_2025}.

To alleviate these limitations, we introduce \emph{Self-Flow-Matching assisted Full Waveform Inversion} (SFM-FWI).
The key distinction is that SFM-FWI does not rely on learning the target model distribution from massive offline datasets.
Instead, it leverages the governing wave physics and the observed measurements to generate \emph{self-supervision} during the inversion. 
As illustrated in Fig.~\ref{fig:diagram}, SFM-FWI alternates between constructing an interpolated intermediate model and training a single flow network online using the FWI data misfit objective, yielding a stable coarse-to-fine model evolution without pretrained priors.

Let \(\mathbf{m}_0\) denote the initial velocity model and \(\hat{\mathbf{m}}_1\) denote the current target estimate, initialized as \(\hat{\mathbf{m}}_1\leftarrow \mathbf{m}_0\) since we do not have ground truth target.
We discretize the outer progression into \(T\) steps and define a normalized time \(t=s/(T-1)\in[0,1]\).
As shown in Algorithm~\ref{alg:sfm_fwi},
at outer step \(s\), we form an interpolated intermediate model
\begin{equation}
\mathbf{m}_t = (1-t)\,\mathbf{m}_0 + t\,\hat{\mathbf{m}}_1.
\label{eq:sfm_interp}
\end{equation}
Given \((\mathbf{m}_t,t)\), the flow network outputs a time-dependent flow velocity field \(\mathbf{v}_{\theta}(\mathbf{m}_t,t)\) in model space.
We then construct a flow-guided proposal for estimating the current target model as
\begin{equation}
\hat{\mathbf{m}}_{1,k} = \mathbf{m}_t + (1-t)\,\mathbf{v}_{\theta}(\mathbf{m}_t,t),
\label{eq:sfm_proposal}
\end{equation}
where \(k=0,\ldots,K-1\) indexes the inner-loop optimization steps of \(\theta\) at the current outer step \(s\).
Note that \(\hat{\mathbf{m}}_{1,k}\) depends on \(\theta\), and \(\mathbf{m}_t\) depends on the current target estimate \(\hat{\mathbf{m}}_1\).
At each step, we update \(\theta\) iteratively by minimizing the FWI data misfit of the \(\hat{\mathbf{m}}_{1,k}\) 
\begin{equation}
L_{t,k}(\theta)= \frac{1}{2}\,\big\|F(\hat{\mathbf{m}}_{1,k})-\mathbf{d}_{\mathrm{obs}}\big\|_2^2,
\label{eq:sfm_loss}
\end{equation}
The gradient with respect to network parameters follows from the chain rule,
\begin{equation}
\nabla_{\theta} L_{t,k}
=
\left(\frac{\partial \hat{\mathbf{m}}_{1,k}}{\partial \theta}\right)^{\!\top}
\left(\nabla_{\hat{\mathbf{m}}_{1,k}} \frac{1}{2}\|F(\hat{\mathbf{m}}_{1,k})-\mathbf{d}_{\mathrm{obs}}\|_2^2\right).
\label{eq:theta_grad_chain}
\end{equation}
Here, \(\frac{\partial \hat{\mathbf{m}}_{1,k}}{\partial \theta}\) is obtained by backpropagation through the flow network and Eq.~\ref{eq:sfm_proposal}.
The term \(\nabla_{\hat{\mathbf{m}}_{1,k}}\frac{1}{2}\|F(\hat{\mathbf{m}}_{1,k})-\mathbf{d}_{\mathrm{obs}}\|_2^2\) is the standard FWI gradient with respect to the model parameters, which is classically computed by the adjoint-state method \citep{Plessix2006}.
In our implementation, we obtain this model gradient via automatic differentiation using the \texttt{Deepwave} differentiable wave-equation solver \citep{richardson_alan_2026}, which enables backpropagation through the time-domain wave-propagation operator \(F(\cdot)\).
This provides an adjoint-equivalent gradient while avoiding manual derivation and implementation of the adjoint equations.

To make the dynamics concrete, we describe the first few outer steps.
At \(s=0\) (i.e., \(t=0\)), the intermediate model is \(\mathbf{m}_t=\mathbf{m}_0\).
The flow network produces an update \(\mathbf{v}_{\theta}(\mathbf{m}_0,0)\), and the proposal becomes \(\hat{\mathbf{m}}_{1,k}=\mathbf{m}_0+\mathbf{v}_{\theta}(\mathbf{m}_0,0)\).
We evaluate the data misfit \( \|F(\hat{\mathbf{m}}_{1,k})-\mathbf{d}_{\mathrm{obs}}\|_2^2 \), compute the corresponding gradient with respect to \(\theta\) and update the \(\theta\) for \(K\) inner steps.
We then set the new target estimate \(\hat{\mathbf{m}}_1 \leftarrow \mathbf{m}_0+\mathbf{v}_{\theta}(\mathbf{m}_0,0)\).

At the next outer step \(s=1\) (small \(t\)), the intermediate state \(\mathbf{m}_t=(1-t)\mathbf{m}_0+t\hat{\mathbf{m}}_1\) remains close to the smooth initial model, so the network input is still dominated by low-wavenumber content.
The flow-guided proposal again minimizes the data misfit, but now the update is conditioned on both the current estimate and the progression time \(t\), gradually introducing finer-scale corrections as \(t\) increases.
This process repeats until \(t\to 1\), and the network learns to refine higher-wavenumber details while maintaining stability.

Standard flow matching trains \(\mathbf{v}_\theta\) by regressing to a prescribed target velocity \((\mathbf{x}_1-\mathbf{x}_0)\) along a chosen path.
In SFM-FWI, we do not have access to paired endpoints \((\mathbf{x}_0,\mathbf{x}_1)\) drawn from a target distribution, nor a ground-truth \(\mathbf{m}_{\mathrm{true}}\) for supervision.
Instead, we use the data-misfit objective to generate \emph{self-supervised} supervision by minimizing Eq.~\ref{eq:sfm_loss}, which encourages the flow proposal \(\hat{\mathbf{m}}_1=\mathbf{m}_t+(1-t)\mathbf{v}_\theta(\mathbf{m}_t,t)\) to move in a direction that reduces waveform residuals.
Across outer steps, the interpolation in Eq.~\ref{eq:sfm_interp} provides a smooth progression from a coarse initial state toward progressively refined estimates, and the shared network \(\mathbf{v}_\theta(\cdot,t)\) learns a time-dependent update rule that consistently decreases the data misfit along this progression.
Empirically, this yields a stable coarse-to-fine transport trajectory that improves both convergence and reconstruction quality, as demonstrated in Section~\ref{sec:experiments}.
Meanwhile, SFM-FWI can be interpreted as learning an update operator that maps a coarse current model to a refined, data-consistent model suggested by short wave-equation-constrained refinements.
As discussed in Section~\ref{sec:discussion}, this induces a deblurring-like effect that enriches wavenumber content in a controlled manner, improving stability in highly nonlinear inversion.
\begin{algorithm}[t]
\caption{Self-flow-matching assisted FWI (SFM-FWI)}
\label{alg:sfm_fwi}
\begin{algorithmic}[1]
\REQUIRE Observed data $\mathbf{d}_{\mathrm{obs}}$; initial model $\mathbf{m}_0$; differential forward modeling operator $F(\cdot)$; outer steps $T$;
inner steps $K$; learning rate $\eta$
\STATE Initialize flow network parameters $\theta$, the target velocity model $\hat{\mathbf{m}}_1\leftarrow\mathbf{m}_0$
\FOR{$s = 0$ \textbf{to} $T-1$}
    \STATE $t=s/(T-1)$
    \STATE Linear interpolation 
    $\mathbf{m}_t = (1 - t) \mathbf{m}_0 + t \hat{\mathbf{m}}_1$
    \FOR{$k = 0$ \textbf{to} $K-1$}
        
        \STATE Predict flow field $\mathbf{v}_{\theta}(\mathbf{m}_t,t)$
        \STATE Proposal $\hat{\mathbf{m}}_{1,k} \leftarrow \mathbf{m}_t + (1-t)\mathbf{v}_{\theta}(\mathbf{m}_t,t)$
        \STATE Loss $L_{t,k} \leftarrow \frac{1}{2}\|F(\hat{\mathbf{m}}_{1,k})-\mathbf{d}_{\mathrm{obs}}\|_2^2$
        \STATE Update $\theta$ using optimizer with learning rate $\eta$
    \ENDFOR
    \STATE Update target model $\hat{\mathbf{m}}_1 \leftarrow \mathbf{m}_t + (1-t)\,\mathbf{v}_{\theta}(\mathbf{m}_t,t)$
\ENDFOR
\RETURN Inverted model $\hat{\mathbf{m}}_{1}$
\end{algorithmic}
\end{algorithm}

\subsection{Neural parameterization of the flow field}
\label{sec:net_details}

We parameterize the time-dependent flow field by a convolutional neural network \(\mathbf{v}_\theta(\mathbf{m}_t,t)\) that maps an intermediate velocity model and normalized time to a model-space update:
\begin{equation}
\mathbf{v}_\theta:\ \mathbb{R}^{H\times W}\times[0,1]\rightarrow \mathbb{R}^{H\times W}.
\end{equation}
We adopt a U-Net to capture multi-scale structures while preserving spatial resolution.
The network outputs an update field with the same spatial dimensions as \(\mathbf{m}_t\), which is used in Eq.~\ref{eq:sfm_proposal} and is optimized online by backpropagating the FWI data misfit in Eq.~\ref{eq:sfm_loss}.
Implementation details are provided in Section~\ref{sec:experiments}.

\section{Numerical Examples}
\label{sec:experiments}
We evaluate the proposed SFM-FWI on three standard synthetic benchmarks (Marmousi \citep{versteeg1994marmousi}, SEG/EAGE Overthrust \citep{aminzadeh1996three}, and a salt body model \citep{billette20052004}) to demonstrate its effectiveness in challenging nonlinear inversion settings.
We compare against three representative \emph{pretraining-free} baselines: (i) conventional least-squares FWI (no explicit regularization),
(ii) FWI with total-variation (TV) regularization,
(iii) Deep Image Prior (DIP)-style reparameterized FWI (DIP-FWI).
We first report results under \emph{clean} synthetic observations for all three benchmarks.
We then present a dedicated robustness study that considers \emph{imperfect observations} (e.g., additive noise and sparse shots) and \emph{poor initial models}, which are common in practice and can severely aggravate cycle skipping.
All methods use the same acoustic wave-equation solver and are tested under identical acquisition geometry, source wavelet, and initialization for fair comparison. In addition, we also match the total number of physics-based optimization steps to ensure comparable computational budgets.
Specifically, SFM-FWI is run with \(T=30\) outer steps and \(K=100\) inner updates per step, while the baseline methods are allocated the same total number of physics-based optimization steps (\(T\times K = 3000\)).

\subsection{Experimental setup}
\label{sec:exp_setup}
\paragraph{Forward modeling and acquisition.}
All experiments are conducted using a 2D constant-density acoustic forward modeling with the software \texttt{Deepwave} \citep{richardson_alan_2026}.
Wave propagation is simulated with a finite-difference time-domain (FDTD) scheme that is second-order accurate in time and eighth-order accurate in space \citep{levander1988fourth}.
Sources are modeled by a Ricker wavelet (with dominant frequency specified in each experiment), and seismograms are generated by sampling the simulated pressure field at receiver locations.
To suppress artificial boundary reflections, we employ perfectly matched layers (PML) \citep{komatitsch2007unsplit} on all boundaries.
We do not impose a free-surface boundary condition at the top. Therefore, free-surface multiples and surface-related reverberations are not included.

\paragraph{Implementation details.}
The initial velocity model is obtained by applying a Gaussian smoothing filter to the ground-truth model, retaining only the long-wavelength background.
But in the test with a poor initial velocity model, we use a linear velocity model as an initial guess.
For all tests using conventional FWI and the TV-regularized one, we enforce simple physical bounds \(c_{\min}\le c\) after each update to prevent nonphysical velocities and improve numerical stability. 
In conventional FWI and conventional FWI with TV regularization, we use an AdamW optimizer \citep{loshchilov2017decoupled} using a learning rate of 2e-4. 
The weight of TV regularization, \(\lambda\), is selected so that both the data loss and the velocity regularization loss are of comparable scale.

The flow field \(\mathbf{v}_\theta(\mathbf{m}_t,t)\) is parameterized by a timestep-conditioned U-Net adapted from the OpenAI guided-diffusion implementation \citep{dhariwal_diffusion_2021}.
The model follows an encoder-decoder structure with skip connections and \(4\) residual blocks per resolution level.
Unless otherwise specified, we set the base channel width to \(64\) and use the default channel multipliers selected by the input velocity size (e.g., for Marmousi sub-region and Salt body: \((1,1,2,2,4,4)\); for other velocity models: \((1,1,2,3,4)\)).
Time conditioning is implemented by a sinusoidal timestep embedding followed by a two-layer MLP with SiLU activations, producing a timestep feature vector of dimension \(4\times 64\). This embedding is injected into every residual block via learned linear projections and additive modulation.
We use GroupNorm normalization and SiLU activations.
The resulting flow network contains approximately 46.62 million parameters for the Marmousi sub-region and Salt body, and 34.70 million parameters for other cases.
The flow network is optimized with AdamW optimizer \citep{loshchilov2017decoupled} using a learning rate of 2e-4.

For DIP-FWI, we parameterize the velocity model as \(\mathbf{m} = g_\phi(\mathbf{z})\), where \(g_\phi\) is a U-Net with the same structure and parameters as that used in SFM-FWI, and \(\mathbf{z}\) is a fixed input (held constant throughout inversion), initialized by the initial velocity models.
The parameters \(\phi\) are optimized by minimizing the FWI data-misfit \(\tfrac{1}{2}\|F(g_\phi(\mathbf{z}))-\mathbf{d}_{\mathrm{obs}}\|_2^2\) with the same optimizer and learning rate as those used in SFM-FWI.
For SFM-FWI and DIP-FWI, we apply a short warm start before waveform-misfit optimization using only the initial model \(\mathbf{m}_0\) as a self-generated reference. Concretely, we minimize an MSE loss that encourages the network output to match \(\mathbf{m}_0\) in both cases:
\begin{equation}
    \mathcal{L}_{\mathrm{warm}}^{\mathrm{SFM}}
=
\|\mathbf{v}_\theta(\mathbf{m}_0,0)-\mathbf{m}_0\|_2^2,
\end{equation}
\begin{equation}
    \mathcal{L}_{\mathrm{warm}}^{\mathrm{DIP}}
=
\|g_\phi(\mathbf{z})-\mathbf{m}_0\|_2^2.
\end{equation}
This warm start is used purely as an implementation heuristic for early numerical stability and consistent network initialization across methods.
It is not used as an additional physical constraint or external prior. 
Empirically, we found this initialization to be slightly more stable than a zero-flow warm start for SFM-FWI.
This warm start incurs negligible additional cost (about 6 seconds in total) relative to the overall inversion time.

Unless otherwise stated, we use the same configuration across all experiments.

\paragraph{Metrics.}
We report qualitative visualizations and quantitative metrics, including the relative L2 error 
\begin{equation}
\mathrm{RelErr}(\mathbf{m},\mathbf{m}_{\mathrm{true}})=\frac{\|\mathbf{m}-\mathbf{m}_{\mathrm{true}}\|_2}{\|\mathbf{m}_{\mathrm{true}}\|_2},
\label{eq:rel_l2}
\end{equation}
and the structural similarity index (SSIM) \citep{wang2004image} measures perceptual and structural agreement between the inverted model \(\mathbf{m}\) and the reference \(\mathbf{m}_{\mathrm{true}}\) by comparing local mean, variance, and cross-covariance. Specifically,
\begin{equation}
\mathrm{SSIM}(\mathbf{m},\mathbf{m}_{\mathrm{true}})
=\frac{\bigl(2\mu_{\mathbf{m}}\mu_{\mathbf{m}_{\mathrm{true}}}+C_1\bigr)\bigl(2\sigma_{\mathbf{m}\,\mathbf{m}_{\mathrm{true}}}+C_2\bigr)}
{\bigl(\mu_{\mathbf{m}}^2+\mu_{\mathbf{m}_{\mathrm{true}}}^2+C_1\bigr)\bigl(\sigma_{\mathbf{m}}^2+\sigma_{\mathbf{m}_{\mathrm{true}}}^2+C_2\bigr)},
\label{eq:ssim}
\end{equation}
where \(\mu_{\mathbf{m}}\) and \(\mu_{\mathbf{m}_{\mathrm{true}}}\) are local means, \(\sigma_{\mathbf{m}}^2\) and \(\sigma_{{\mathbf{m}}_{\mathrm{true}}}^2\) are local variances, and \(\sigma_{{\mathbf{m}}\,{\mathbf{m}}_{\mathrm{true}}}\) is the local cross-covariance between \({\mathbf{m}}\) and \({\mathbf{m}}_{\mathrm{true}}\). In practice, these local statistics are estimated using a Gaussian-weighted window. The constants \(C_1\) and \(C_2\) are included for numerical stability, typically defined as \(C_1=(K_1L)^2\) and \(C_2=(K_2L)^2\), where \(L\) is the dynamic range of the signal and \(K_1, K_2\) are small scalars (e.g., \(K_1=0.01\), \(K_2=0.03\)). To avoid instability, the SSIM score is calculated on the normalized velocity models.

\subsection{Marmousi}
The Marmousi model is a widely used benchmark characterized by strong lateral velocity variations and complex layered structures.
These features make gradient-based inversion particularly challenging from a smooth initialization, where cycle skipping and illumination imbalance can degrade inversion performance, especially at depth.
To provide a clear validation-and-analysis pipeline, we first conduct experiments on a cropped Marmousi sub-region, and then test on the whole modified Marmousi model in a more realistic acquisition setting.

\begin{figure}[!htp]
    \centering
    \includegraphics[width=1.0\linewidth]{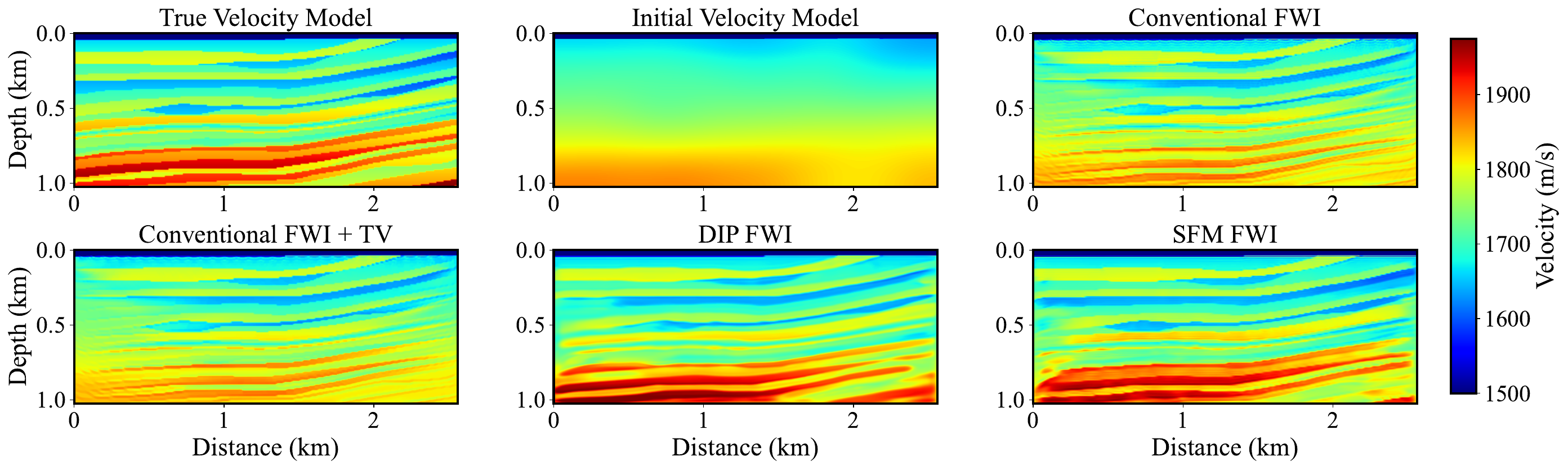}
    \caption{Marmousi sub-region inversion results (clean data).
    Top row: true velocity model, Gaussian-smoothed initial model, and conventional FWI result.
    Bottom row: TV-regularized FWI, DIP-FWI, and the proposed SFM-FWI.}
    \label{fig:fwi_marmousi_part}
\end{figure}

\paragraph{Marmousi sub-region.}
The cropped Marmousi sub-region used in this experiment is shown in Fig.~\ref{fig:fwi_marmousi_part}.
The model spans the area of $2.56 \times 1.02$ $\mathrm{km}^2$, discretized on a regular grid with the horizontal and vertical intervals of 4 $\mathrm{m}$.
31 shots are uniformly placed on the surface, with 106 receivers per shot placed on the surface as well.
The recording length is 1.6 seconds with a time interval of 4 $\mathrm{ms}$.
Sources are injected as Ricker wavelets with a dominant frequency of 25 Hz. 
The TV regularization weight \(\lambda\) is set as 1e-10.

\begin{figure}
    \centering
    \includegraphics[width=1.0\linewidth]{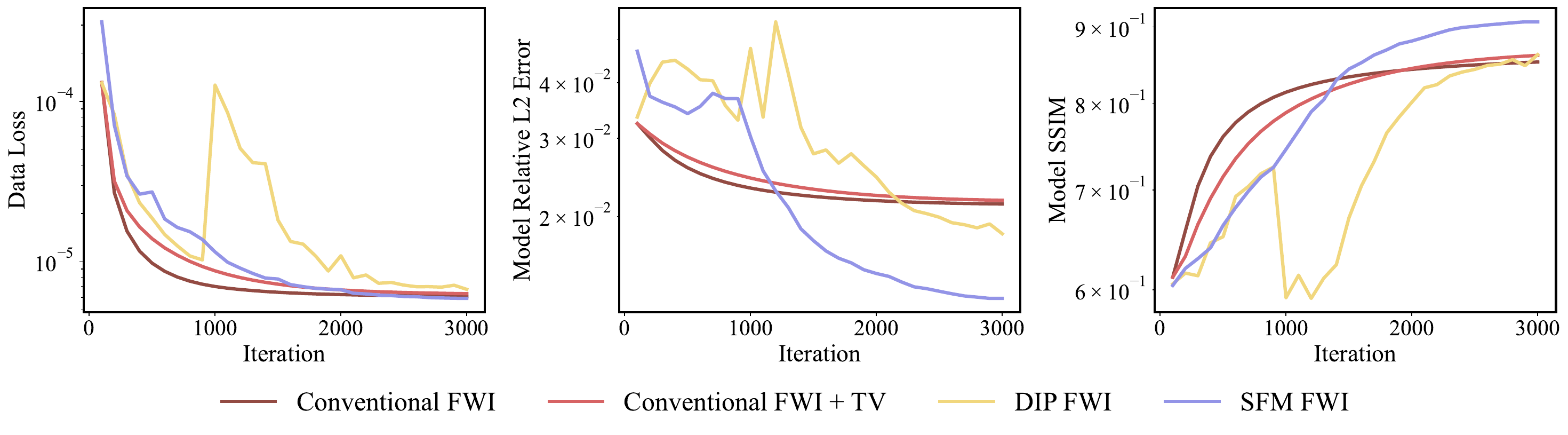}
    \caption{Convergence curves on the Marmousi sub-region benchmark. Left: data-misfit versus iteration. Middle: relative L2 error of the inverted velocity model with respect to the ground truth. Right: SSIM versus iteration.}
    \label{fig:fwi_marmousi_subregion_convergence_curve}
\end{figure}

We perform SFM-FWI and other baseline approaches on the same initial velocity model. 
We report convergence curves in Fig.~\ref{fig:fwi_marmousi_subregion_convergence_curve}, including the data misfit, relative L2 error, and SSIM versus iteration.
SFM-FWI reduces the data misfit to a lower final value, while consistently achieving lower model error and higher SSIM than the baselines.
The inverted results are shown in Fig.~\ref{fig:fwi_marmousi_part}.
Starting from a smoothed initial velocity model, conventional FWI is able to recover major layered features in the shallow-to-mid depths.
However, the recovered model remains noticeably over-smoothed and exhibits reduced resolution at depth. Thin layering and small-scale lateral variations are only partially reconstructed even with high-frequency data, and the deeper portion of the model shows weaker structural continuity, consistent with limited illumination and the tendency of gradient-based updates to be dominated by well-illuminated shallow regions.
Introducing TV regularization improves the sharpness of several interfaces and suppresses small-scale oscillatory artifacts.
Nevertheless, the TV-regularized result displays a characteristic piecewise-smooth bias. Continuous velocity gradients are artificially homogenized, and certain stratified features exhibit discretized, block-like characteristics, which can misrepresent the inherently gradual variations present in the Marmousi background model and diminish the geological realism in regions with smoothly varying properties.
The DIP-FWI produces visually sharper textures and richer high-frequency details than conventional FWI.
At the same time, it shows a noticeable amplitude bias, particularly in deeper regions, where the recovered velocity trend appears over-amplified compared with the ground truth.
In contrast, SFM-FWI provides the best overall trade-off between resolution and physical plausibility.
It reconstructs sharper reflectors and improved layer continuity across the model while maintaining a background velocity trend closer to the ground truth.
Most importantly, in the deeper part of the model where illumination is weaker, SFM-FWI preserves coherent stratification and recovers more reliable intermediate-to-high wavenumber content than the baselines.
These results indicate that the proposed flow-guided updates help stabilize the inversion trajectory and mitigate depth-dependent resolution loss, yielding models that are more consistent with both wave-equation data fitting and plausible subsurface structure. 
Table~\ref{tab:model_comparison_paritial_marmousi} quantitatively confirms the visual observations in Fig.~\ref{fig:fwi_marmousi_part}.
SFM-FWI achieves the lowest relative L2 error and the highest SSIM among all methods, indicating improved accuracy and structural consistency with the ground truth.
\begin{table}[!htp]
\centering
\caption{Quantitative comparison of different approaches on Marmousi sub-region.}
\label{tab:model_comparison_paritial_marmousi}
\begin{tabular}{lcc}
\toprule
\textbf{Approaches} & \textbf{Relative L2 error}$\downarrow$ & \textbf{SSIM}$\uparrow$   \\
\midrule
Conventional FWI & 0.0213 & 0.853  \\
Conventional FWI with TV & 0.0217 & 0.862  \\
DIP-FWI & 0.0177 & 0.866  \\
SFM-FWI & \textbf{0.0130} & \textbf{0.907}  \\
\bottomrule
\end{tabular}
\end{table}

\begin{figure}[!htp]
    \centering
    \includegraphics[width=1.0\linewidth]{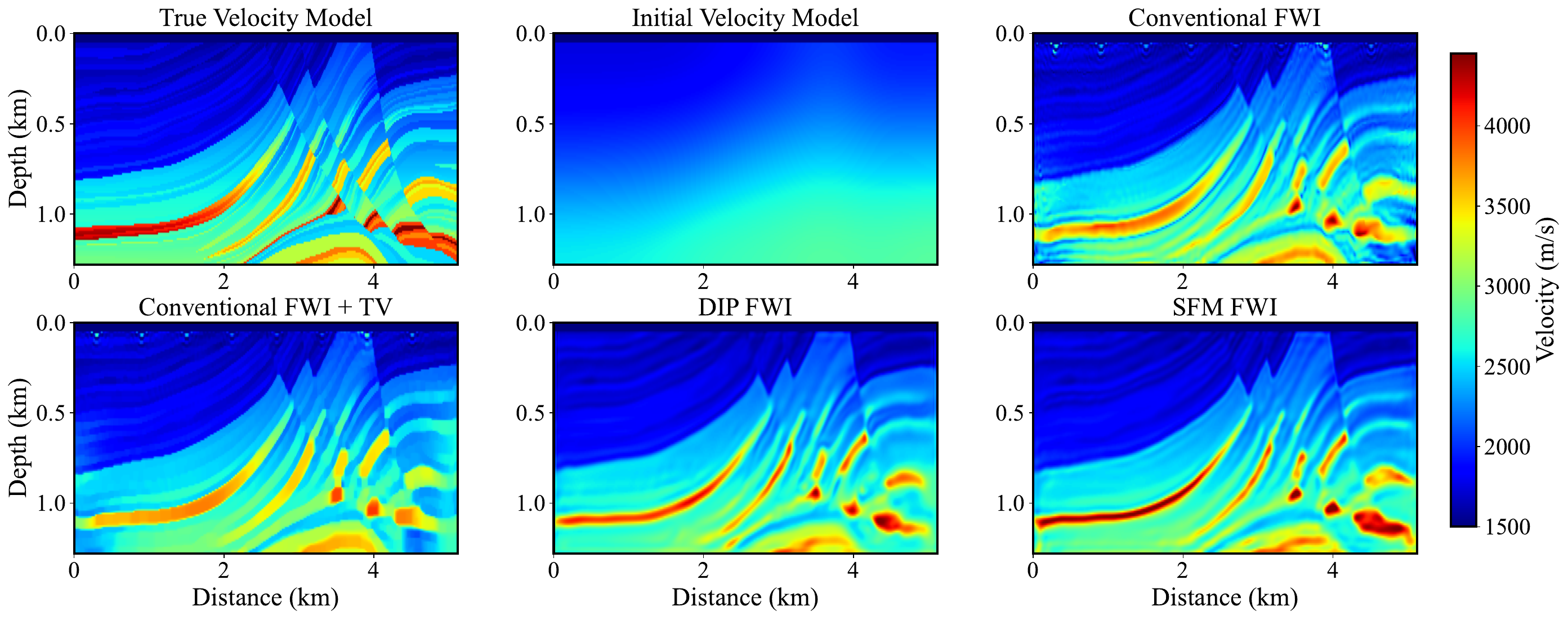}
    \caption{Marmousi inversion results (clean data).
    Top row: true velocity model, Gaussian-smoothed initial model, and conventional FWI result.
    Bottom row: TV-regularized FWI, DIP-assisted FWI, and the proposed SFM-FWI.}
    \label{fig:fwi_marmousi}
\end{figure}

\paragraph{More realistic-scale test.}
We then evaluate all methods on a larger modified Marmousi model to assess performance at a more realistic scale (Fig.~\ref{fig:fwi_marmousi}).
The model spans an area of \(5.12\times1.28~\mathrm{km}^2\) and is discretized with a uniform grid spacing of \(10~\mathrm{m}\).
We use \(25\) shots and \(85\) receivers per shot, uniformly distributed along the surface.
The source wavelet is a Ricker wavelet with a dominant frequency of \(8~\mathrm{Hz}\).
Seismic data are recorded for \(2.4~\mathrm{s}\) with a time sampling interval of \(\Delta t=4~\mathrm{ms}\).
The TV regularization weight \(\lambda\) is set as 1e-8.

Fig.~\ref{fig:fwi_marmousi} shows the inversion comparison of different approaches, and quantitative analysis is shown in Table~\ref{tab:model_comparison_marmousi}. 
SFM-FWI consistently outperforms all baseline methods with fewer artifacts, and the deeper part shows more accurate reconstruction of the velocity models.
\begin{table}[!htp]
\centering
\caption{Quantitative comparison of different approaches on Marmousi.}
\label{tab:model_comparison_marmousi}
\begin{tabular}{lcc}
\toprule
\textbf{Approaches} & \textbf{Relative L2 error}$\downarrow$ & \textbf{SSIM}$\uparrow$   \\
\midrule
Conventional FWI & 0.0933 & 0.762  \\
Conventional FWI with TV & 0.0934 & 0.786  \\
DIP-FWI & 0.0765 & 0.779  \\
SFM-FWI & \textbf{0.0682} & \textbf{0.791}  \\
\bottomrule
\end{tabular}
\end{table}

\subsection{Overthrust}
We next consider the Overthrust benchmark, which is characterized by thrust-related structures, steeply dipping reflectors, and strong velocity contrasts.
These features amplify the nonlinearity of the waveform misfit and often cause conventional FWI to converge to suboptimal solutions, especially in regions with complex wavepaths and reduced illumination (e.g., beneath the thrust zone).
The model spans the area of $11.52\times 1.92$ $\mathrm{km}^2$ with a spatial interval of 25 m.
We simulate 31 shots with sources uniformly placed on the surface. 
We recorded the data with 128 receivers, and the data were generated using a Ricker wavelet with a dominant frequency of 15 Hz, a recording length of 4s, and a sampling rate of 2ms.
The TV regularization weight \(\lambda\) is set as 1e-8.

\begin{figure}[!htp]
    \centering
    \includegraphics[width=1.0\linewidth]{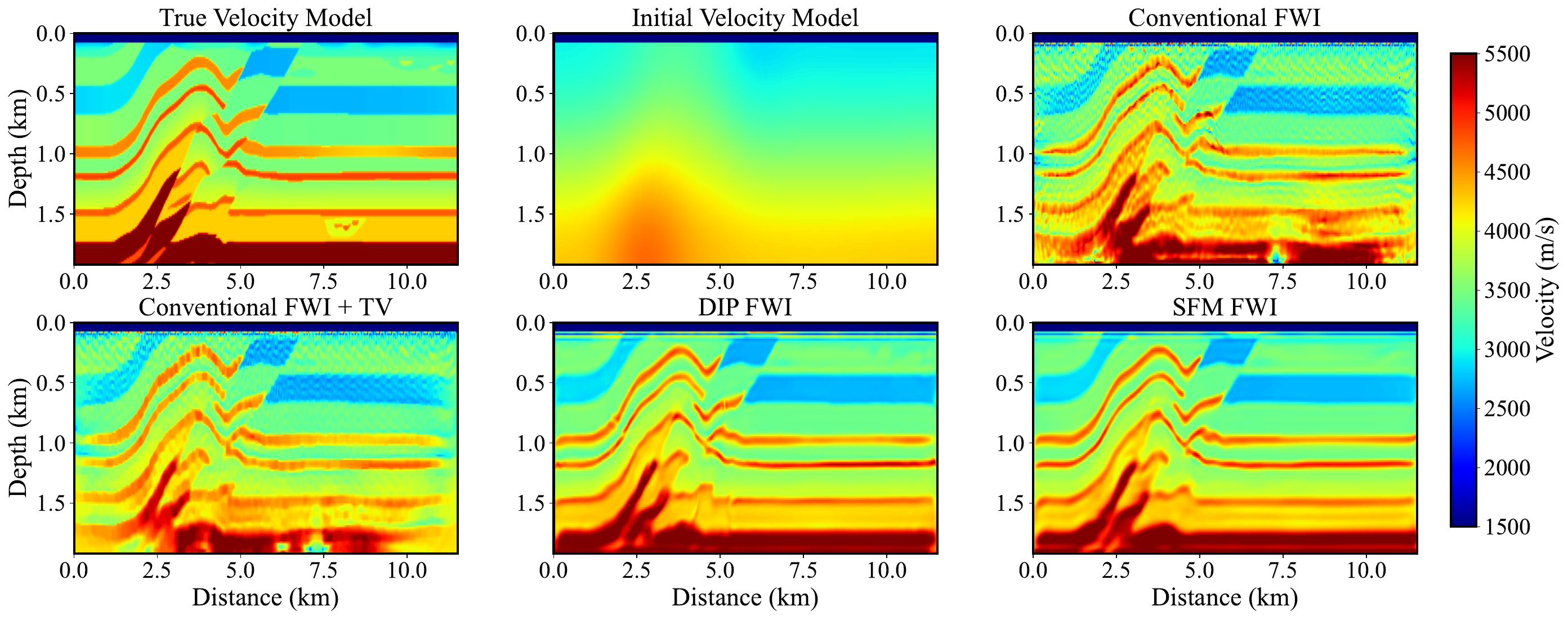}
    \caption{Overthrust inversion results (clean data).
    Top row: true velocity model, Gaussian-smoothed initial model, and conventional FWI result.
    Bottom row: TV-regularized FWI, DIP-assisted FWI, and the proposed SFM-FWI.}
    \label{fig:fwi_overthrust}
\end{figure}

The inversion results are shown in Fig.~\ref{fig:fwi_overthrust}.
Conventional FWI recovers the main large-scale trend and portions of the thrust-related layering, but exhibits pronounced illumination-related artifacts and high-frequency striping, particularly in the deeper part of the model and near the complex thrust zone.
TV regularization suppresses oscillatory artifacts and sharpens several interfaces, yet it still suffers from residual striping and tends to bias the model toward piecewise-smoothed structures.
DIP-FWI reconstructs sharper reflectors and enhances fine-scale texture compared with conventional FWI, but it also shows noticeable localized inconsistencies near complex structures.
In contrast, SFM-FWI produces the most geologically consistent reconstruction with reduced artifacts and improved continuity of steeply dipping reflectors across the thrust region.
Notably, the deeper part of the model is reconstructed more stably. 
These results suggest that the proposed flow-guided updates provide a robust mechanism to balance resolution enhancement and stability under severe nonlinearity and illumination imbalance in Overthrust-style settings.
Table~\ref{tab:model_comparison_overthrust} provides quantitative results on the Overthrust benchmark.
Consistent with the visual comparison in Fig.~\ref{fig:fwi_overthrust}, SFM-FWI achieves the best overall performance, attaining the lowest relative L2 error (0.0390) and the highest SSIM (0.869).
Compared with conventional FWI, SFM-FWI reduces the relative error by nearly 60\% and improves SSIM by more than 0.20, indicating substantially improved structural fidelity in this strongly nonlinear setting.
\begin{table}[!htp]
\centering
\caption{Quantitative comparison of different approaches on Overthrust.}
\label{tab:model_comparison_overthrust}
\begin{tabular}{lcc}
\toprule
\textbf{Approaches} & \textbf{Relative L2 error}$\downarrow$ & \textbf{SSIM}$\uparrow$   \\
\midrule
Conventional FWI & 0.0962 & 0.664  \\
Conventional FWI with TV & 0.0959 & 0.742  \\
DIP-FWI & 0.0443 & 0.851  \\
SFM-FWI & \textbf{0.0390} & \textbf{0.869}  \\
\bottomrule
\end{tabular}
\end{table}

\subsection{Salt body}
We finally consider a salt-body benchmark, which is widely regarded as one of the most challenging settings for waveform-based inversion due to the strong velocity contrast between salt and surrounding sediments.
Such high-contrast inclusions induce complex wave phenomena (strong reflections and refractions) and often create poor illumination beneath salt, where small boundary mispositioning can trigger severe cycle skipping and lead to spurious updates.
The velocity model spans an area of $7.68\times3.84$ $\mathrm{km}^2$ with a spatial interval of 15 $\mathrm{m}$.
There are 42 shots and 168 receivers uniformly placed on the surface. Observed data are generated using a Ricker wavelet with dominant frequency 10 Hz, with a total length of 4 $\mathrm{s}$ and a sampling rate of 2 $\mathrm{ms}$. The TV regularization weight \(\lambda\) is set as 1e-8.

\begin{figure}[!htp]
    \centering
    \includegraphics[width=1.0\linewidth]{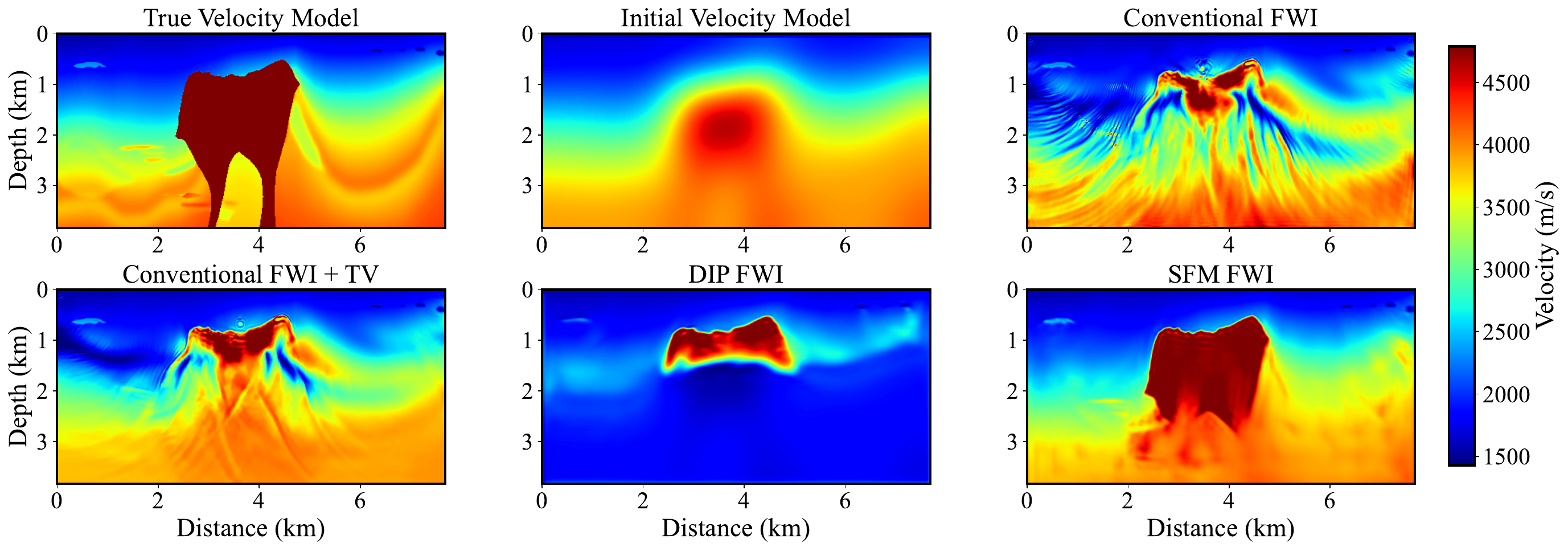}
    \caption{Salt body inversion results (clean data).
    Top row: true velocity model, Gaussian-smoothed initial model, and conventional FWI result.
    Bottom row: TV-regularized FWI, DIP-assisted FWI, and the proposed SFM-FWI.}
    \label{fig:fwi_bp}
\end{figure}

\begin{figure}[!htp]
    \centering
    \includegraphics[width=1.0\linewidth]{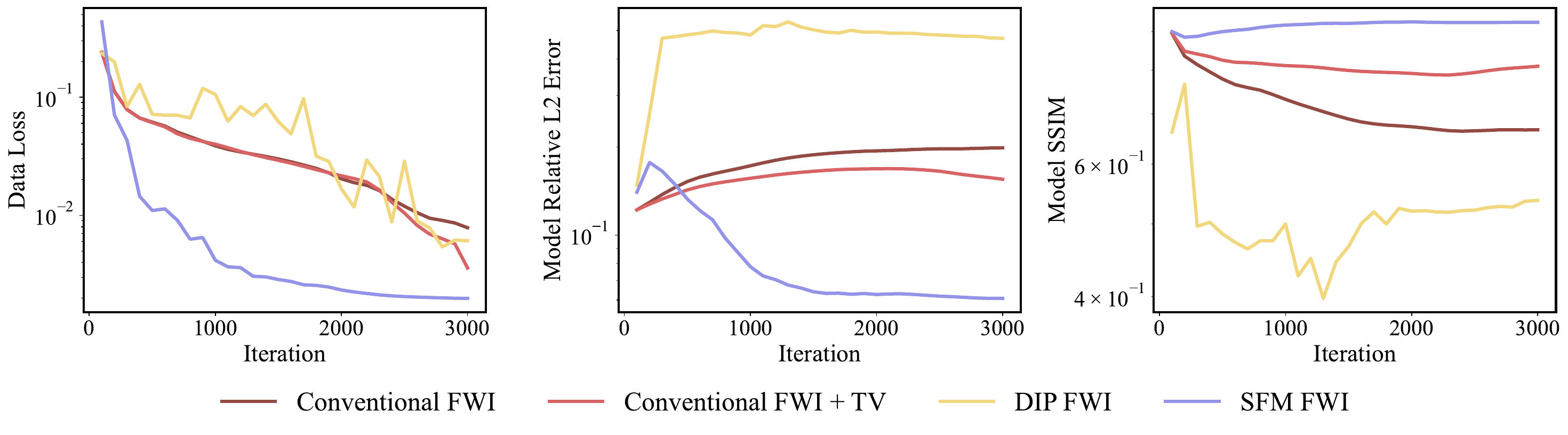}
    \caption{Convergence curves on the salt-body benchmark. Left: data-misfit versus iteration. Middle: relative L2 error of the inverted velocity model with respect to the ground truth. Right: SSIM versus iteration.}
    \label{fig:fwi_bp_convergence_curve}
\end{figure}

We applied these four approaches, and the corresponding inverted results are shown in Fig.~\ref{fig:fwi_bp}.
The conventional FWI is unstable around the salt body, while it partially introduces a high-velocity region near the top of the salt, the recovered salt geometry is distorted, and the sub-salt region is contaminated by strong coherent artifacts, consistent with illumination loss and cycle skipping in high-contrast settings.
TV regularization suppresses oscillatory components and improves boundary sharpness, but still produces noticeable distortions around the salt flanks and residual sub-salt artifacts.
Although DIP-FWI delineates the top-of-salt interface reasonably well in our salt experiment, its overall model error is substantially worse than the other approaches.
This behavior is consistent with two compounding factors.
First, salt-body inversion is highly ill-posed due to the extreme velocity contrast and the resulting illumination imbalance. The strong reflections and refractions at the top of salt create a sub-salt shadow zone where the data provide weak constraints on deeper structures, and small boundary mispositioning can trigger severe nonlinearity and cycle skipping.
Second, DIP-FWI relies on implicit regularization from the network parameterization and its optimization dynamics.
Under complex geology, the fixed-input DIP-FWI formulation can produce artifacts and unstable reconstructions.
Without time progression, DIP has no mechanism to control wavenumber introduction rate, making it vulnerable to overfitting in weak-data regions.
In contrast, SFM-FWI provides a substantially more faithful reconstruction of both the salt body and the surrounding sediments.
The salt boundaries are sharper and better positioned, the salt interior is more uniform, and sub-salt artifacts are markedly reduced compared to all baselines.
Overall, SFM-FWI yields the most geologically plausible model and the most stable reconstruction behavior in this high-contrast regime.
Table~\ref{tab:model_comparison_bp} confirms these observations.
SFM-FWI achieves the lowest relative L2 error (0.060) and the highest SSIM (0.926), significantly outperforming other baselines.
As shown in Fig.~\ref{fig:fwi_bp_convergence_curve}, we further report the convergence behavior of all methods on the salt-body benchmark in terms of the data misfit, relative L2 model error, and SSIM.
SFM-FWI rapidly reduces data misfit in the early iterations and continues to decrease to the lowest final value among all approaches, indicating a more stable and effective optimization trajectory.
This improved data fitting is accompanied by consistently better model quality, with the relative L2 error decreasing monotonically to its minimum and the SSIM remaining the highest throughout the inversion.
In contrast, conventional FWI and TV-regularized FWI exhibit slower misfit reduction and plateau at higher error levels, while DIP-FWI shows unstable behavior with elevated model error and lower SSIM, consistent with its degraded reconstruction performance in this high-contrast, weak-illumination setting.
These results demonstrate that the proposed flow-guided, physics-driven updates provide robust stabilization and superior recovery of salt geometry and sub-salt velocity characterization.
\begin{table}[!htp]
\centering
\caption{Quantitative comparison of different approaches on Salt body.}
\label{tab:model_comparison_bp}
\begin{tabular}{lcc}
\toprule
\textbf{Approaches} & \textbf{Relative L2 error}$\downarrow$ & \textbf{SSIM}$\uparrow$   \\
\midrule
Conventional FWI & 0.1983 & 0.667  \\
Conventional FWI with TV & 0.1534 & 0.811  \\
DIP-FWI & 0.4736 & 0.531  \\
SFM-FWI & \textbf{0.0607} & \textbf{0.926}  \\
\bottomrule
\end{tabular}
\end{table}

\subsection{Robustness study}
\label{sec:robustness}
In practice, FWI performance is strongly affected by imperfect observations and acquisition limitations, and is particularly sensitive to the quality of the starting model.
To assess practical robustness beyond the clean-data benchmarks, we further evaluate robustness on Marmousi under three stress scenarios that frequently arise in practice: (i) poor initialization, (ii) noisy observations, and (iii) sparse shot acquisition.
All methods are evaluated under the same forward modeling setup and acquisition setting as used in the full Marmousi test, except for the sparse shots scenario, where we reduce the shots from 25 shots to 5 shots.
Quantitative results are summarized in Table~\ref{tab:model_comparison_marmousi_various}.

\begin{figure}[!htp]
    \centering
    \includegraphics[width=1.0\linewidth]{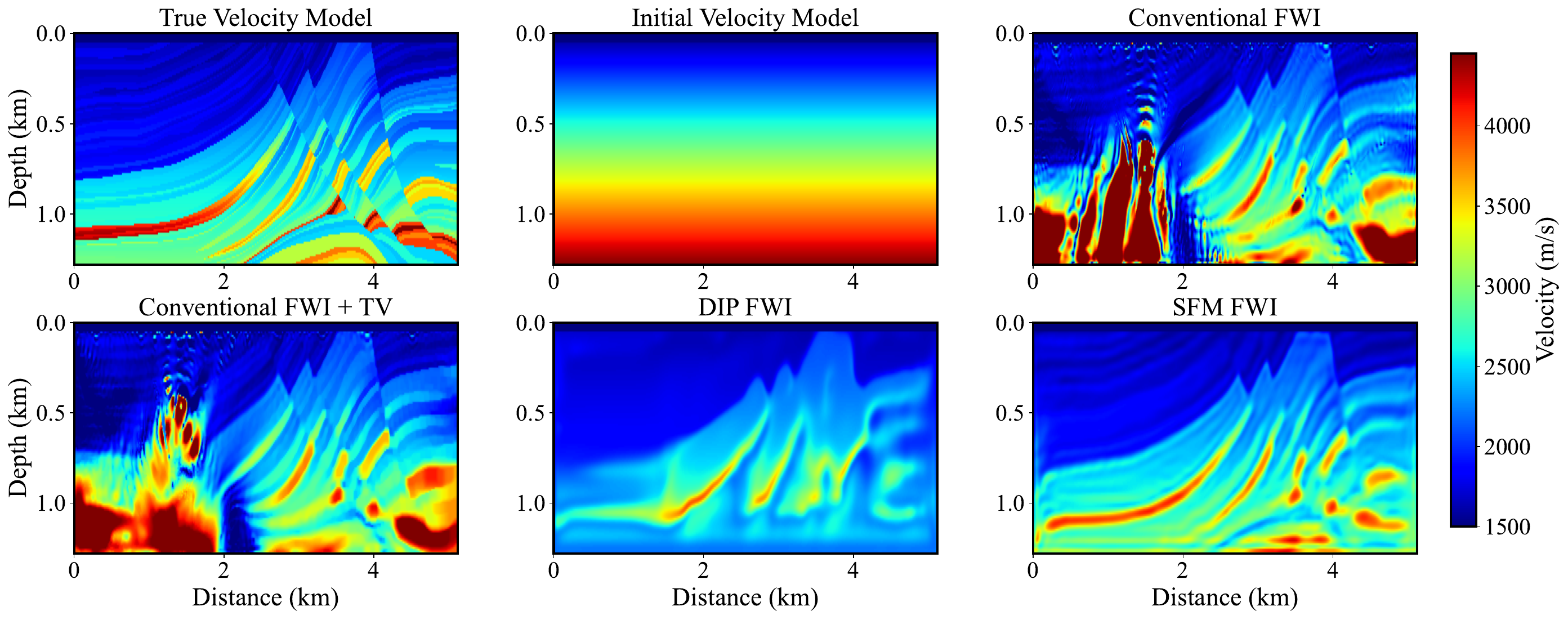}
    \caption{Marmousi inversion results (poor initial model).
    Top row: true velocity model, Gaussian-smoothed initial model, and conventional FWI result.
    Bottom row: TV-regularized FWI, DIP-assisted FWI, and the proposed SFM-FWI.}
    \label{fig:fwi_marmousi_poor_initial}
\end{figure}

\paragraph{Poor initial velocity models.}
To emulate inaccurate background models, we construct degraded initial velocities by directly using a linear increased velocity model.
Such initializations typically increase the risk of cycle skipping because predicted and observed waveforms can become misaligned by more than half a period.
The TV regularization weight \(\lambda\) is set as 1e-8.
As shown in Table~\ref{tab:model_comparison_marmousi_various} and Fig.~\ref{fig:fwi_marmousi_poor_initial}, conventional FWI degrades substantially under this setting.
TV regularization improves stability but remains far from satisfactory, reflecting its tendency to enforce piecewise-smooth structure without fully resolving long-wavelength misalignment.
DIP-FWI offers a better background improvement, yet still exhibits notable residual errors in the inverted results.
In contrast, SFM-FWI remains stable inversion and achieves the best accuracy and SSIM, indicating that the proposed flow-guided updates effectively maintain a coarse-to-fine reconstruction trajectory even when the starting model is severely degraded.

\begin{figure}[!htp]
    \centering
    \includegraphics[width=0.7\linewidth]{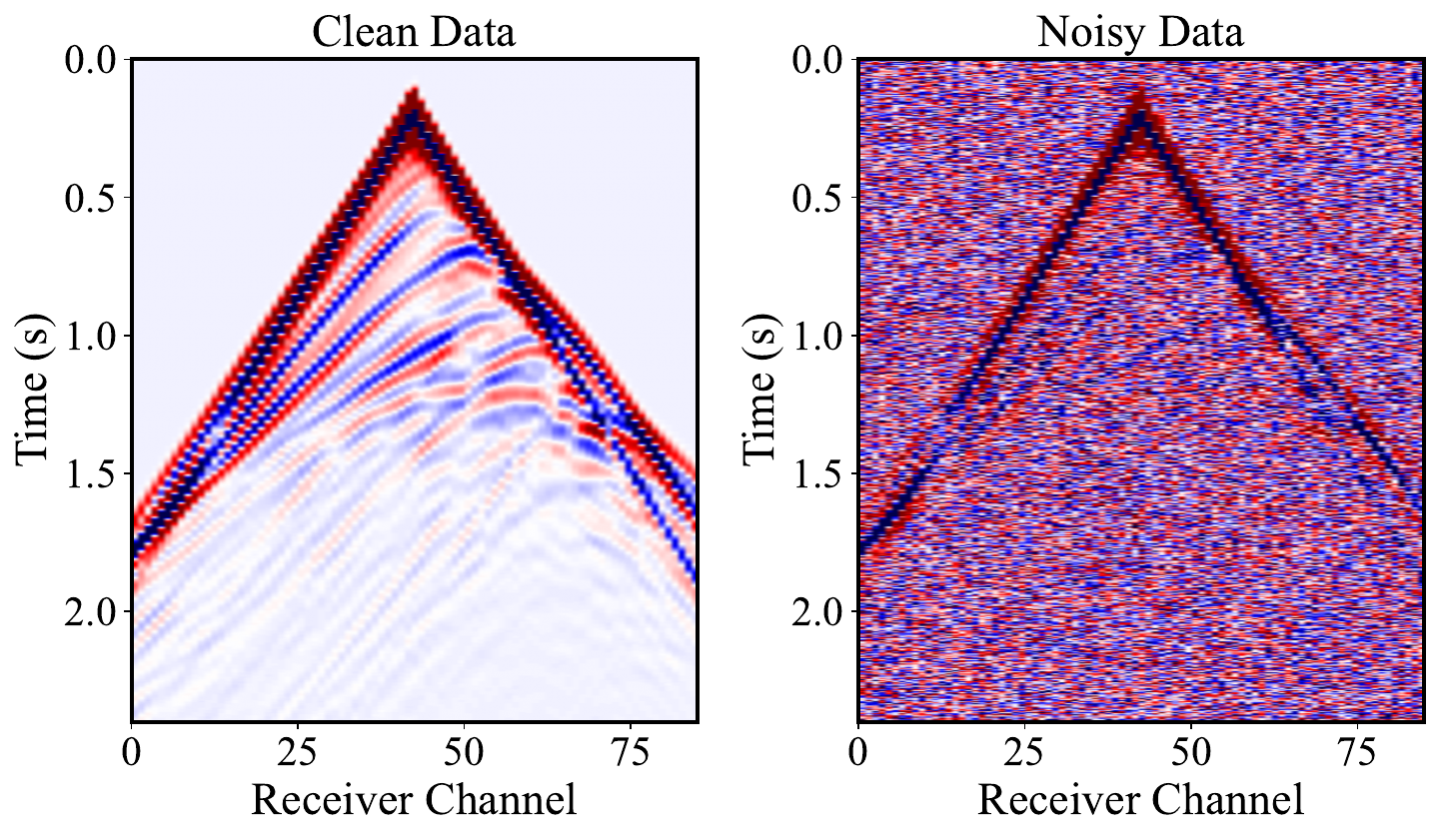}
    \caption{Simulated synthetic shot gathers recorded at 2.5 km (left) before adding noise, and corresponding noisy data (right).}
    \label{fig:marmousi_noisy_data}
\end{figure}

\begin{figure}[!htp]
    \centering
    \includegraphics[width=1.0\linewidth]{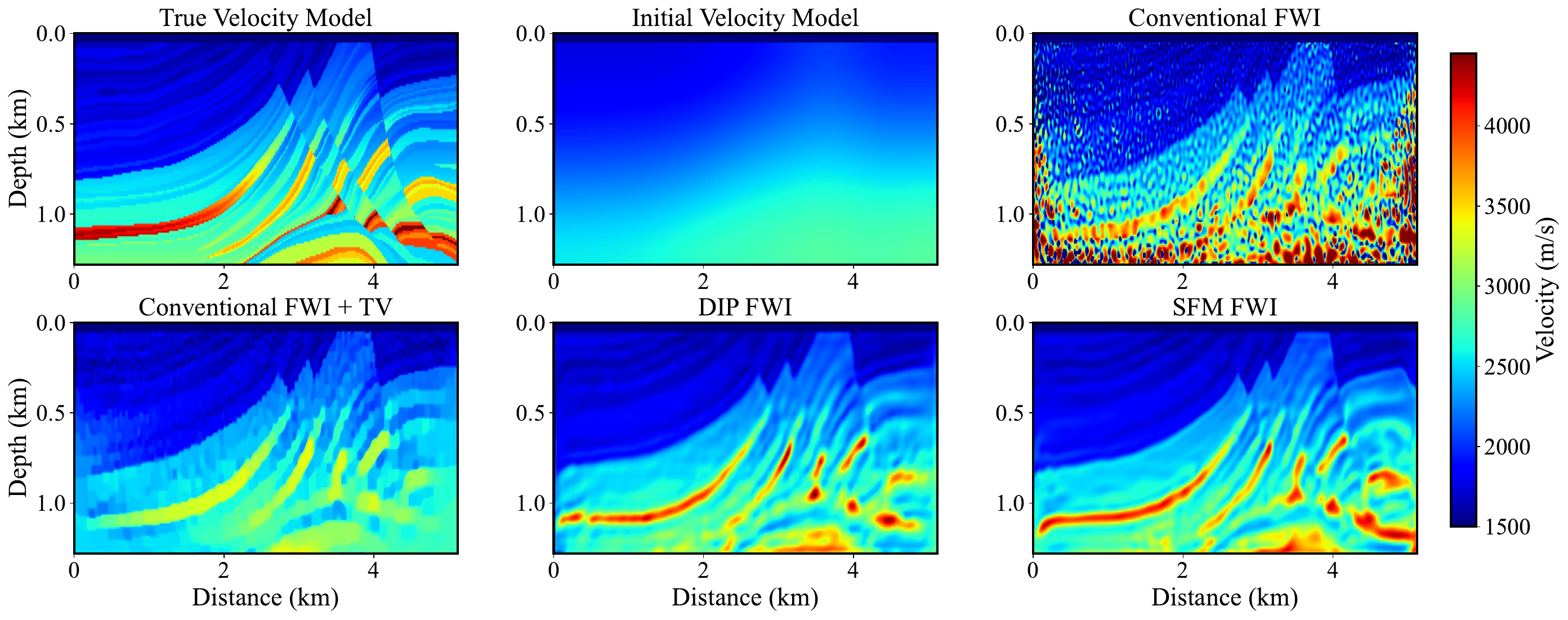}
    \caption{Marmousi inversion results (noisy data).
    Top row: true velocity model, Gaussian-smoothed initial model, and conventional FWI result.
    Bottom row: TV-regularized FWI, DIP-assisted FWI, and the proposed SFM-FWI.}
    \label{fig:fwi_marmousi_noised}
\end{figure}

\paragraph{Noisy observations.}
We further test robustness to imperfect measurements by contaminating the synthetic data with additive Gaussian noise at a signal-to-noise ratio of 3.5 dB, as shown in Fig.~\ref{fig:marmousi_noisy_data}.
The TV regularization weight \(\lambda\) is set as 1e-7.
In this case, we reduce the total iterations to 1500 for all approaches.
As shown in Table~\ref{tab:model_comparison_marmousi_various} and Fig.~\ref{fig:fwi_marmousi_noised}, in the noisy setting, conventional FWI experiences strong performance deterioration and exhibits artifacts in the inversion, suggesting that waveform fitting becomes unstable when the residual is dominated by noise.
TV regularization and DIP-FWI are notably more robust, consistent with their ability to suppress noise-like components, but TV regularization demonstrates an obvious piecewise phenomenon.
SFM-FWI attains the best overall performance, achieving the highest SSIM and the lowest error, which indicates improved preservation of coherent structures and reduced sensitivity to noisy residuals.

\begin{figure}[!htp]
    \centering
    \includegraphics[width=1.0\linewidth]{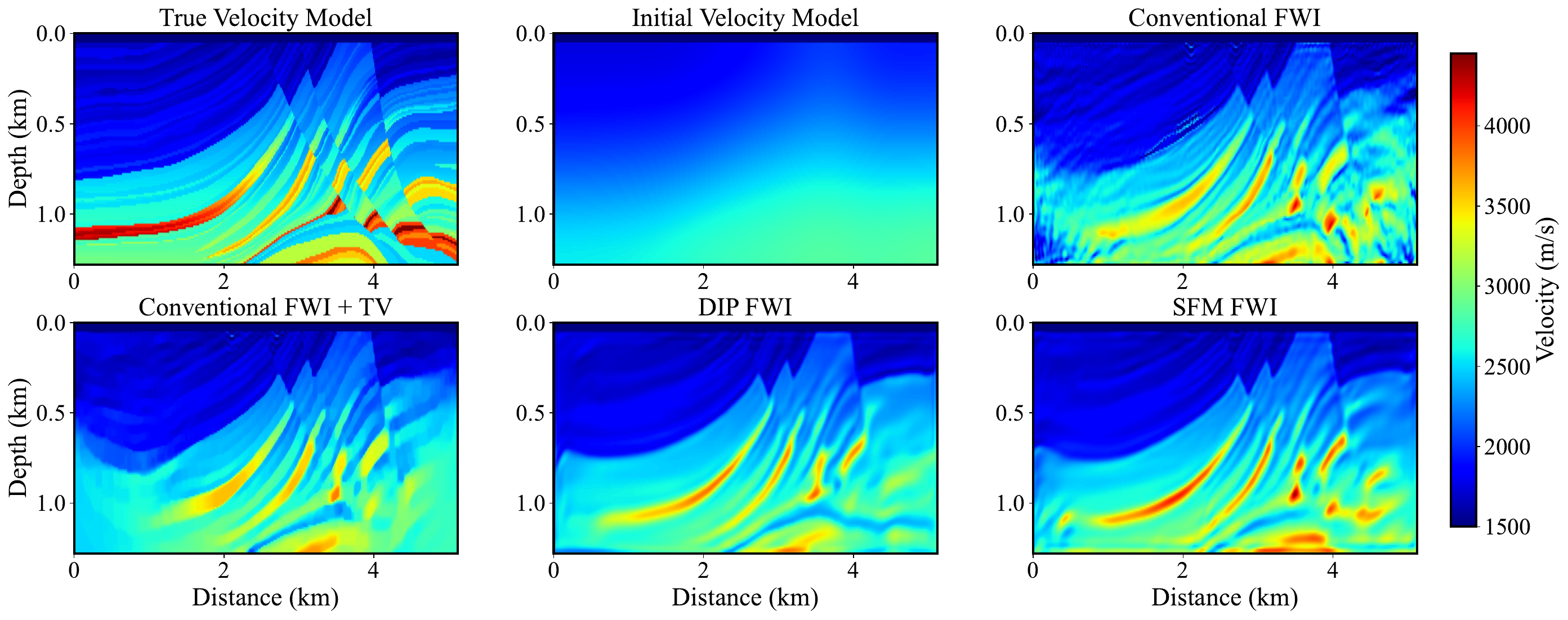}
    \caption{Marmousi inversion results (sparse shots).
    Top row: true velocity model, Gaussian-smoothed initial model, and conventional FWI result.
    Bottom row: TV-regularized FWI, DIP-assisted FWI, and the proposed SFM-FWI.}
    \label{fig:fwi_marmousi_sparse_shots}
\end{figure}

\paragraph{Sparse shot acquisition.}
Finally, we evaluate all methods under reduced acquisition by subsampling shots, as sparse acquisition reduces illumination and coverage, which typically leads to depth-dependent resolution loss and acquisition-footprint artifacts.
We place only 5 shots uniformly on the surface.
The TV regularization weight \(\lambda\) is set as 3e-8.
In this case, to avoid network overfitting, we reduce the U-Net to 2 residual blocks per resolution, resulting in a model with 21.42 million parameters.
Under the sparse-shot setting, all methods show performance degradation relative to the clean-data case, and the differences among baselines become smaller (Table \ref{tab:model_comparison_marmousi_various}), but we could see a clear acquisition footprint in conventional FWI and that with TV regularization (Fig.~\ref{fig:fwi_marmousi_sparse_shots}).
Nevertheless, SFM-FWI continues to deliver the best quantitative results, outperforming conventional FWI and providing consistent gains over TV and DIP baselines.
This suggests that the proposed framework remains effective under limited acquisition by stabilizing updates through an online learned transport field while remaining anchored to wave-equation data fitting.

\paragraph{Summary.}
Overall, Table~\ref{tab:model_comparison_marmousi_various} demonstrates that SFM-FWI provides consistent robustness improvements across all three stress scenarios.
The largest gain is observed for poor initialization, where SFM-FWI significantly mitigates cycle-skipping-induced failure modes, while also maintaining strong performance under noisy observations and reduced acquisition.
\begin{table*}[t]
\centering
\caption{Quantitative comparison of different approaches on Marmousi under three scenarios.}
\label{tab:model_comparison_marmousi_various}
\setlength{\tabcolsep}{2.pt}
\begin{tabular}{lcc|cc|cc}
\toprule
\multirow{2}{*}{\textbf{Approach}} 
& \multicolumn{2}{c|}{\textbf{Poor initialization}} 
& \multicolumn{2}{c|}{\textbf{Noisy observations}} 
& \multicolumn{2}{c}{\textbf{Sparse shots}} \\
\cmidrule(lr){2-3}\cmidrule(lr){4-5}\cmidrule(lr){6-7}
& \textbf{Relative L2 error}$\downarrow$ & \textbf{SSIM}$\uparrow$
& \textbf{Relative L2 error}$\downarrow$ & \textbf{SSIM}$\uparrow$
& \textbf{Relative L2 error}$\downarrow$ & \textbf{SSIM}$\uparrow$ \\
\midrule
FWI    &   0.4234  &  0.494    &  0.2219  &  0.299  &  0.1292  &  0.639  \\
FWI + TV   & 0.2766 &  0.525   &  0.1298  &  0.663  &  0.1249  &  0.688  \\
DIP-FWI           &    0.1813   &  0.596   &  0.0943  &  0.711  &  0.1160  &  0.692  \\
SFM-FWI                 &  \textbf{0.0835}  &  \textbf{0.732}  &  \textbf{0.0857}  &  \textbf{0.731}  &  \textbf{0.1059}  &  \textbf{0.712}  \\
\bottomrule
\end{tabular}
\end{table*}

\section{Discussion}
\label{sec:discussion}

The numerical results on Marmousi, Overthrust, and salt benchmarks, together with the robustness study under poor initialization, noisy observations, and sparse acquisition, consistently indicate that SFM-FWI improves reconstruction fidelity and stability over conventional FWI and representative regularized baselines.
In this section, we provide additional insights into \emph{why} the proposed approach behaves favorably.
Specifically, we analyze the evolution of model complexity during inversion, taking the Marmousi sub-region as an example, illustrate the emergence of an automatic layer-stripping behavior, discuss connections to deblurring, and summarize practical considerations such as computational cost and compatibility with pretrained generative priors.
In addition, we perform ablation study and additional comparison with diffusion regularized FWI.

\subsection{Automatic coarse-to-fine (multi-scale) inversion}
\begin{figure}[!htp]
    \centering
    \includegraphics[width=0.6\linewidth]{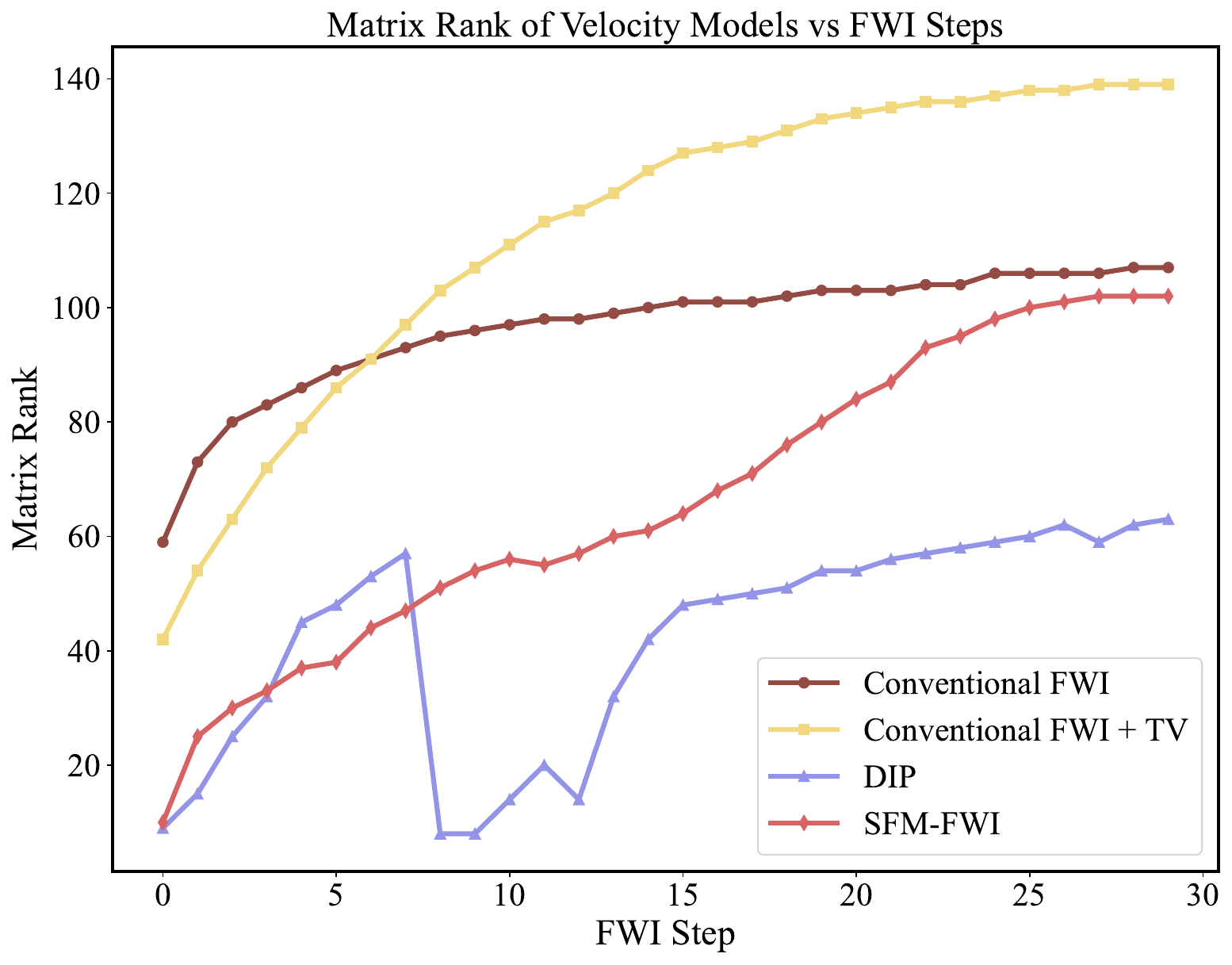}
    \caption{The rank evolution of inverted velocity models with different approaches.}
    \label{fig:rank_curve}
\end{figure}

To better understand the implicit regularization behavior of different methods, we analyze the evolution of the model rank during inversion (Fig.~\ref{fig:rank_curve}).
We obtain the effective rank of the velocity model \(\mathbf{m}\) by performing Singular Value Decomposition (SVD) on the model matrix with a shape of $n_x\times n_z$. 
The rank is determined by the number of singular values $\sigma_i$ that are significantly larger than zero, using a numerical tolerance $\epsilon$. In our implementation, we follow the standard convention where $\epsilon = \sigma_{max} \cdot \max(n_x, n_z) \cdot \eta$, where $\eta$ is the machine precision. This ensures that the rank captures the essential structural complexity of the reconstructed velocity field while being robust to noise.
Low-rank models are dominated by smooth, long-wavelength components, while higher-rank models indicate progressively richer fine-scale structures.
A common failure mode in strongly nonlinear FWI is the premature introduction of high-wavenumber components, which can amplify cycle skipping and trap the inversion in poor local minima.
Compared with conventional FWI and baseline regularization strategies, SFM-FWI exhibits a noticeably slower and more monotone increase of rank across iterations.
This indicates that SFM-FWI naturally prioritizes the recovery of low-wavenumber background structures and gradually transitions to higher-wavenumber refinement, effectively realizing an \emph{automatic multi-scale inversion schedule} without handcrafted frequency continuation.
Such a coarse-to-fine evolution is particularly beneficial under limited illumination and noisy observations, where stabilizing the early-stage updates is critical for avoiding cycle skipping.

\subsection{Layer stripping from inversion evolution}
\begin{figure}[!htp]
    \centering
    \includegraphics[width=1.0\linewidth]{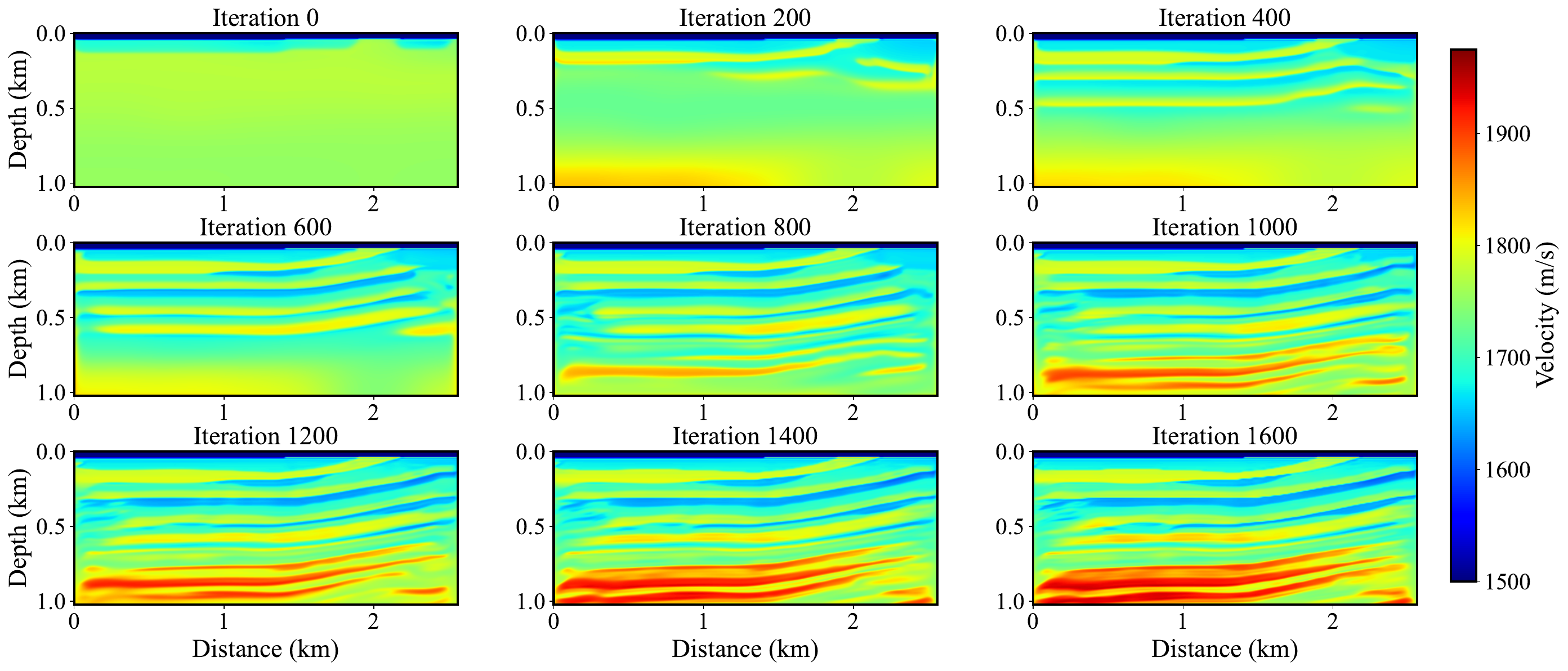}
    \caption{The inversion evolution at different iteration steps.}
    \label{fig:fwi_partial_marmousi_sfmfwi_evolution}
\end{figure}

In addition to the rank evolution, we visualize intermediate velocity models throughout the inversion process, as shown in Fig.~\ref{fig:fwi_partial_marmousi_sfmfwi_evolution}.
Interestingly, SFM-FWI exhibits a clear \emph{layer-stripping} behavior, in which large-scale background and shallow structures are recovered first, and deeper interfaces and finer details are progressively revealed as iterations proceed.
This behavior is favorable for FWI, especially when the initial velocity model is poor in quality. 
More importantly, this layer stripping emerges \emph{automatically} without any handcrafted design and without the need for hyperparameter selection.
In addition, SFM-FWI tends to prevent premature injection of high-wavenumber components that can destabilize the inversion.
It helps stabilize the inversion under strong nonlinearity and illumination imbalance, improves the interpretability of intermediate models, and ultimately enables more reliable recovery of deep structures beneath complex formations.

\subsection{Connection to deblurring}
\label{sec:deblurring}

\begin{figure}[!htp]
    \centering
    \includegraphics[width=1.0\linewidth]{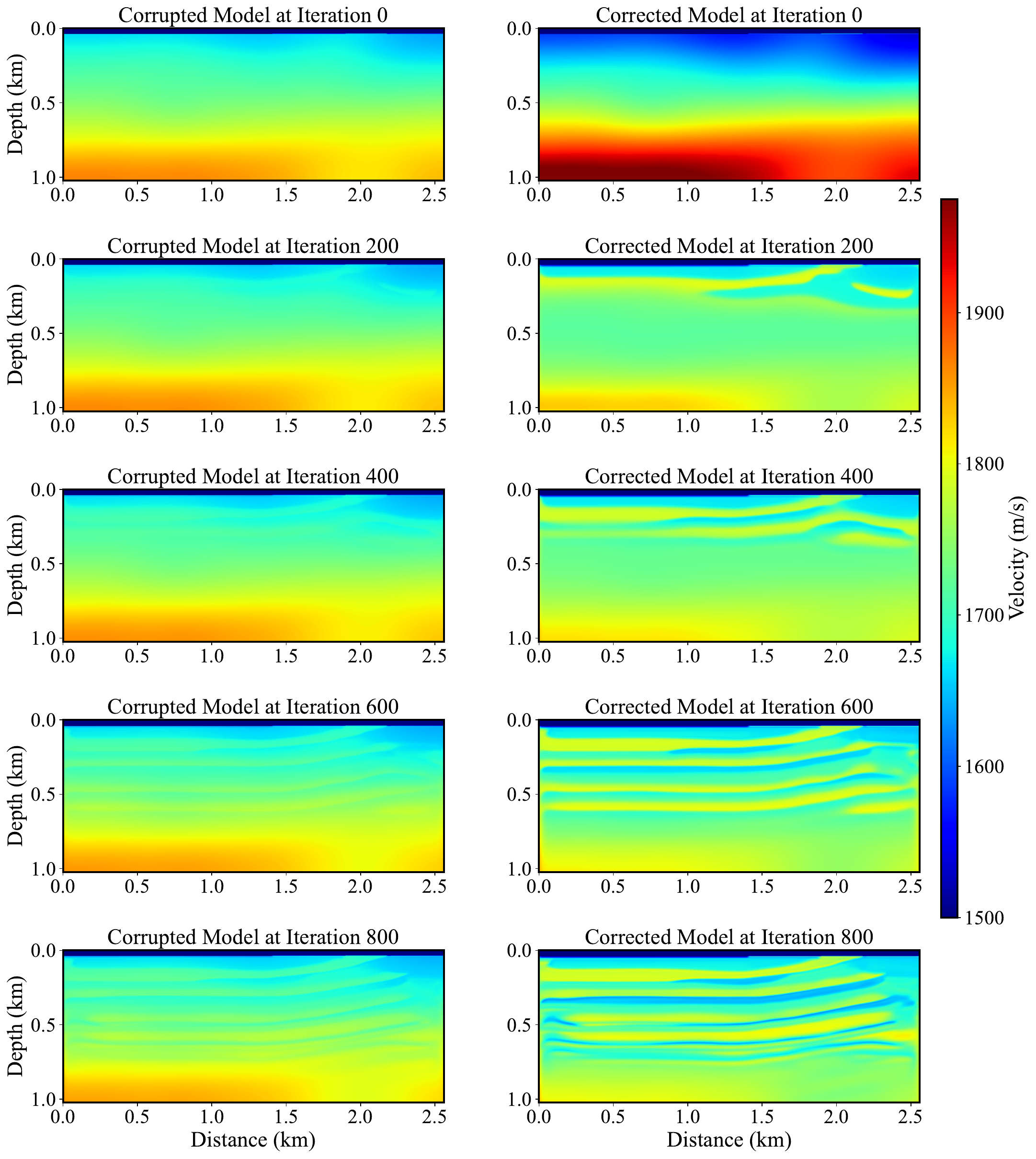}
    \caption{Potential deblurring (inverse-Hessian-like) effect of SFM-FWI. The left panel represents the corrupted velocity models, and the right panel represents the corresponding corrected velocity models after the flow network prediction.}
    \label{fig:fwi_partial_marmousi_sfmfwi_evolution_deblurring}
\end{figure}

We next provide quantitative evidence supporting a deblurring-like interpretation of the flow-guided update.
As shown in Fig.~\ref{fig:fwi_partial_marmousi_sfmfwi_evolution_deblurring}, the interpolated intermediate model (corrupted input) is biased toward the smoothed initialization and thus dominated by low-wavenumber content, whereas the corrected output after the flow update exhibits visibly sharper interfaces and richer layering.

To quantify these observations, we compare radially averaged Fourier spectra and gradient-based sharpness metrics between the corrupted and corrected models at iterations (0, 200, 400, 600, 800).
Let \(\mathbf{m}_{\mathrm{corrupt}}\) be the corrupted intermediate model and \(\mathbf{m}_{\mathrm{corr}}\) the corrected model after the flow update.
\(R_{\mathrm{band}}\) denotes the ratio of band-limited radial spectral content (corrected/corrupted) over \(k\in[0.03,0.10]\) cycles/m.
HF gain is the ratio of the high-wavenumber fraction beyond \(k_c=0.0375\) cycles/m (corrected/corrupted).
\(\|\nabla m\|^2\) gain is the ratio of mean gradient energy \(\mathbb{E}\|\nabla m\|^2\) (corrected/corrupted), and \(p90(\|\nabla m\|)\) gain is the ratio of the 90th-percentile gradient magnitude (corrected/corrupted). 
We report these metrics in Table~\ref{tab:deblur_metrics}.

In the mid-wavenumber band \(k\in[0.03,0.10]\) cycles/m, the corrected model consistently contains more spectral content than the corrupted input, with the band-limited ratio increasing from 1.33 at iteration 0 to 1.73 at iteration 800.
Moreover, spatial sharpness increases markedly: the mean gradient energy \(\mathbb{E}\|\nabla m\|^2\) increases by \(3.03\times\) and the 90th-percentile gradient magnitude increases by \(3.78\times\) at iteration 800, indicating substantially sharper and more coherent interfaces.
These results show that the flow-guided update does not merely amplify arbitrary high-frequency noise; instead, it progressively introduces structured, interface-aligned detail in a controlled coarse-to-fine manner under physical guidance from the observed data.
While we do not claim a formal second-order equivalence, this measured sharpening effect is consistent with an inverse-Hessian-like (deblurring) interpretation at an empirical level.
\begin{table}[t]
\centering
\caption{Quantitative evidence of the deblurring-like effect. }
\label{tab:deblur_metrics}
\begin{tabular}{ccccc}
\toprule
Iter. & \(R_{\mathrm{band}}\uparrow\) & HF gain & \(\|\nabla m\|_2^2\) gain & \(p90(\|\nabla m\|)\) gain \\
\midrule
0   & 1.33 & 0.74 & 0.46 & 2.07 \\
200 & 1.22 & 0.91 & 1.15 & 3.14 \\
400 & 1.45 & 1.07 & 1.43 & 5.02 \\
600 & 1.60 & 1.12 & 2.33 & 4.43 \\
800 & 1.73 & 1.16 & 3.03 & 3.78 \\
\bottomrule
\end{tabular}
\end{table}

\subsection{Additional Computational Cost}
\label{sec:cost}
A practical advantage of SFM-FWI is that it incurs minimal additional computational overhead.
The dominant cost in all waveform inversion methods remains wave-equation solves and adjoint computations.
In SFM-FWI, the training of the flow network introduces additional forward/backward passes through a U-Net, but these costs are small compared to PDE solves.
We report the total runtime per inversion run (minutes) in Table~\ref{tab:cost} on an NVIDIA H200. All runs use the same forward solver and matched physics-step budgets within each benchmark.
Under our default settings, SFM-FWI achieves improved reconstruction quality with nearly the same number of physics-based evaluations as conventional FWI baselines, and the wall-clock time increase is marginal in practice.

\begin{table}[t]
\centering
\caption{Computational cost comparison (wall-clock time) per inversion run (minutes). }
\label{tab:cost}
\begin{tabular}{lcccc}
\toprule
\textbf{Benchmark} & \textbf{FWI} & \textbf{FWI+TV} & \textbf{DIP-FWI} & \textbf{SFM-FWI} \\
\midrule
Marmousi Sub-region & 20.85 & 20.89 & 23.36 & 23.38  \\
Marmousi        & 9.01 & 8.99 & 10.96 & 10.93 \\
Overthrust       & 35.50 & 35.18 & 37.39 & 37.57 \\
Salt body            & 47.17 & 47.09 & 50.31 & 50.33 \\
\bottomrule
\end{tabular}
\end{table}

\subsection{Compatibility with pretrained generative priors}
\label{sec:combine_pretrained}
Although SFM-FWI is designed to avoid reliance on offline datasets, it is also compatible with pretrained generative priors when such models are available.
For example, a pretrained diffusion or flow model can be used to initialize the network parameters or provide additional regularization signals, after which SFM-FWI can refine the transport field online to adapt to the target survey and observed measurements.
This suggests a hybrid future direction where pretrained priors provide broad geological knowledge, and SFM-FWI provides physics-driven adaptation and stabilization, potentially improving robustness under distribution shift.

\subsection{Ablation: sensitivity to the number of outer and inner steps}
\label{sec:ablation_TK}

\begin{figure}[!htp]
    \centering
    \includegraphics[width=1.0\linewidth]{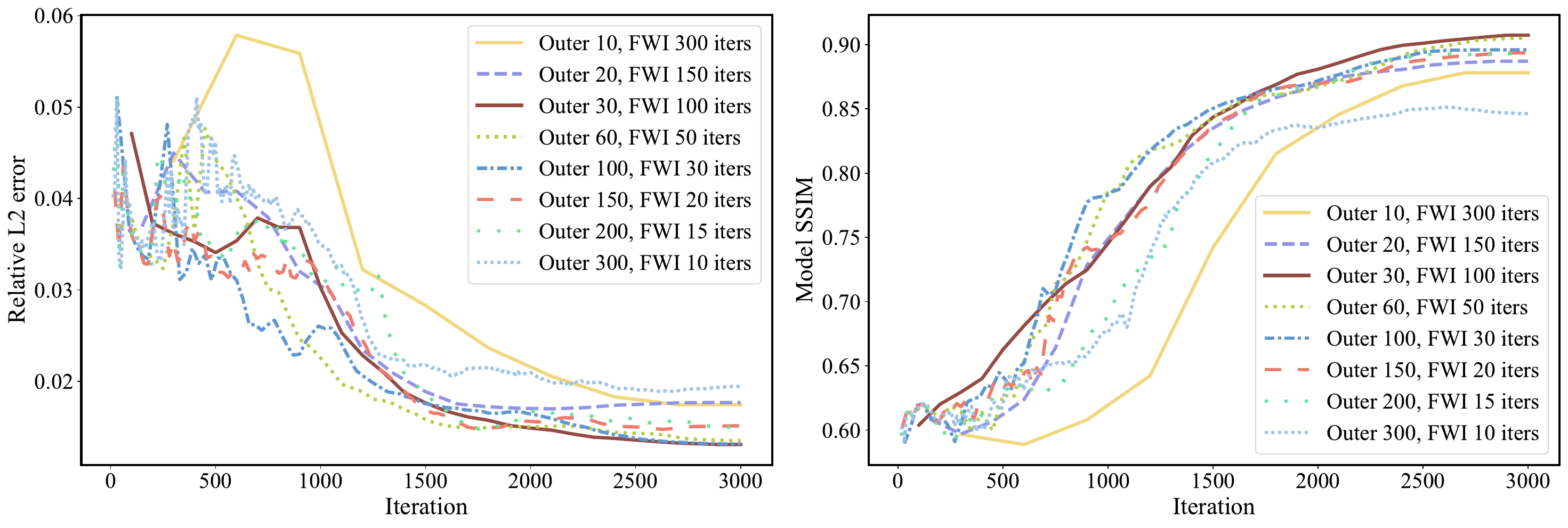}
    \caption{Ablation study on the sensitivity of SFM-FWI to the number of outer steps \(T\) and inner steps \(K\) on the Marmousi sub-region benchmark under a fixed budget \(T\times K=3000\).
    Left: relative L2 error versus iteration. Right: SSIM versus iteration.}
    \label{fig:ablation_TK}
\end{figure}

We study the sensitivity of SFM-FWI to the number of outer flow steps \(T\) and inner FWI steps \(K\) on the Marmousi sub-region benchmark.
Recall that each outer step performs \(K\) inner updates of the flow network using the FWI data-misfit, followed by updating the target model.
To provide practical guidance under a fixed computational budget, we keep the total number of inner updates constant, i.e., \(T\times K = 3000\), and vary \((T,K)\) across \((10,300)\), \((20,150)\), \((30,100)\), \((60,50)\), \((100,30)\), \((150,20)\), \((200,15)\), \((300,10)\).
Fig.~\ref{fig:ablation_TK} reports the convergence curves of relative L2 error and SSIM.

Overall, SFM-FWI is robust to moderate changes in \((T,K)\): most configurations with \(T\in[20,200]\) achieve similar final accuracy and SSIM.
However, extreme choices degrade performance.
Using very few outer steps (e.g., \(T=10\)) results in slower convergence and inferior final metrics, suggesting that insufficient outer progression limits the coarse-to-fine evolution.
Conversely, using too many outer steps with very small inner loops (e.g., \(T=300, K=10\)) also underperforms, indicating that overly short inner refinements provide weak training signals for the flow field and can lead to suboptimal updates.

Across tested settings, \(T=30\) and \(K=100\) provides the best trade-off between convergence speed and final reconstruction quality on this benchmark, and we adopt this setting as the default in the rest of the paper.
More broadly, our results suggest that selecting \(T\) in a moderate range (tens to a few hundreds) and allocating sufficient inner refinement steps (tens to \(\sim 10^2\)) yields stable performance, while keeping the total budget \(T\times K\) aligned with the available compute.

\subsection{Comparison with a pretrained diffusion-prior baseline}
\label{sec:discussion_diffusion}

We further conduct a direct comparison to a diffusion-prior-assisted FWI baseline for a comprehensive study to highlight a complementary advantage of SFM-FWI under prior mismatch.

We use the released DiffusionFWI weights from \citep{wang_prior_2023}, pretrained on the OpenFWI dataset, and apply them to the Overthrust sub-region benchmark without finetuning.
We emphasize that the DiffusionFWI baseline is tested on a \(0.64\times0.64~\mathrm{km}^2\) sub-region because the available pretrained model is trained on \(64\times64\) inputs.
We generate data using a 15~Hz Ricker wavelet with a 1~ms time step and a total recording length of 1.5~s.
There are 8 shots and 32 receivers placed on the surface over a \(0.64\times0.64~\mathrm{km}^2\) domain.
All methods share the same computational budget and initialization for fair comparison.

\begin{figure}[!htp]
    \centering
    \includegraphics[width=1.0\linewidth]{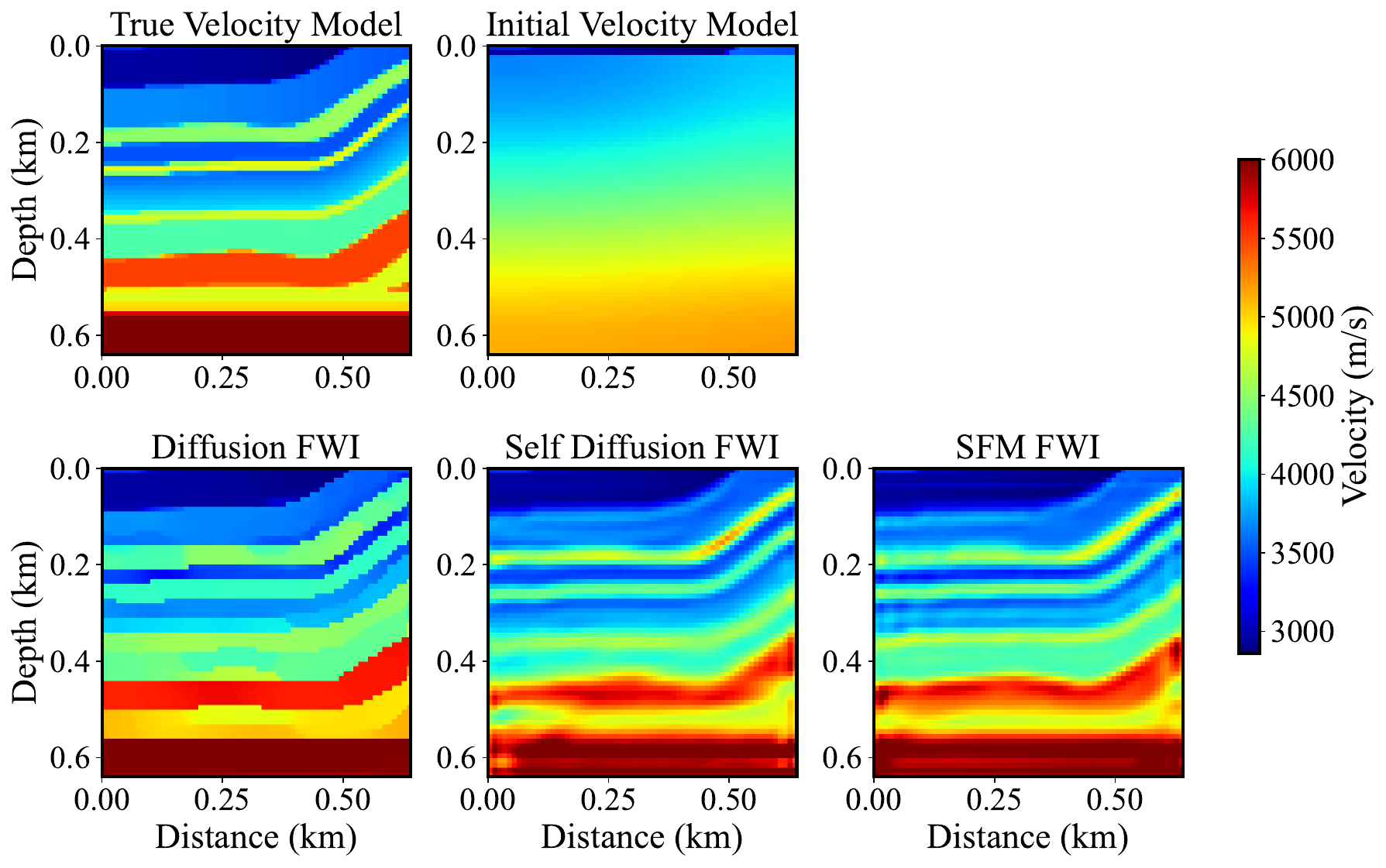}
    \caption{Comparison with a pretrained diffusion-prior baseline on the Overthrust benchmark.
    From left to right: true velocity model, smoothed initial model, DiffusionFWI using pretrained weights from \citep{wang_prior_2023} (trained on OpenFWI), a pretraining-free self-diffusion variant of our approach (Self-diffusion FWI), and the proposed SFM-FWI.
    The relative L2 errors of those approaches are 0.0530, 0.0445, and 0.0365, while the corresponding SSIMs are 0.7806, 0.8336, and 0.8730. 
    SFM-FWI better matches both structure and amplitudes, yielding the lowest model error and the highest SSIM.}
    \label{fig:diffusion_baseline}
\end{figure}

As shown in Fig.~\ref{fig:diffusion_baseline}, DiffusionFWI recovers a visually plausible structural layout, indicating that the pretrained diffusion prior encodes useful spatial patterns.
However, we observe a clear amplitude mismatch: the recovered velocity tends to exhibit piecewise-constant characteristics, whereas the Overthrust model contains vertical transitions.
This is consistent with the distribution shift between the OpenFWI training distribution (predominantly piecewise-layered models) and the target Overthrust setting, which biases the inversion toward the training-set morphology.

In contrast, SFM-FWI learns its flow field \emph{online} from the wave-equation data misfit and therefore adapts to the target observation and physics without relying on an external velocity-model database.
On this benchmark, as shown in Table~\ref{tab:diffusion_mismatch}, SFM-FWI achieves lower model error and higher SSIM than DiffusionFWI, supporting our main motivation that physics-driven self-supervision can mitigate prior mismatch.
To isolate the effect of the \emph{starting distribution}, we additionally evaluate a self-diffusion variant of our framework that follows a diffusion-style generative trajectory initialized from random noise, while still using the same wave-equation data-misfit for self-supervision.
This variant removes reliance on offline pretrained datasets, but reintroduces a key practical drawback of diffusion sampling in the FWI setting, where the inversion must traverse a long denoising trajectory from a noise-like state, which can slow convergence and reduce stability under highly nonlinear wave-physics constraints.
Quantitatively, the self-diffusion variant attains Relative L2 error \(0.0445\) and SSIM \(0.8336\), improving over DiffusionFWI under prior mismatch (Relative L2 error \(0.0530\), SSIM \(0.7806\)) but remaining worse than SFM-FWI.
\begin{table}[t]
\centering
\caption{Quantitative comparison under prior mismatch on the Overthrust sub-region.}
\label{tab:diffusion_mismatch}
\begin{tabular}{lcc}
\toprule
\textbf{Method} & \textbf{Relative L2 error}$\downarrow$ & \textbf{SSIM}$\uparrow$ \\
\midrule
DiffusionFWI (pretrained on OpenFWI) & 0.0530 & 0.7806 \\
Self-diffusion FWI (random-noise start)  & 0.0445 & 0.8336 \\
SFM-FWI (online, physics-driven)     & \textbf{0.0365} & \textbf{0.8730} \\
\bottomrule
\end{tabular}
\end{table}

\subsection{Future work}
\label{sec:limitations}

Although effective, this current study has several limitations. 
First, all experiments are conducted in a 2D acoustic and constant-density setting.
Extending SFM-FWI to elastic/anisotropic physics and to 3D acquisition is an important next step. 
Second, we do not include tests on full-field data, where additional modeling errors due to source signature uncertainty, attenuation, and anisotropy, may further challenge inversion. 
Incorporating field-data preprocessing and more realistic physics is left for future work.
Third, our diffusion-prior comparison is limited to a single small Overthrust sub-region (\(0.64\times0.64~\mathrm{km}^2\)) for a comprehensive study, as the available pretrained DiffusionFWI model is trained on a \(64\times64\) input resolution. 
Extending diffusion baselines to larger domains via retraining or patch-based inference is an interesting direction for future study.

\section{Conclusion}
\label{sec:conclusion}

We proposed Self-Flow-Matching assisted Full Waveform Inversion (SFM-FWI), a physics-driven framework that integrates flow matching into wave-equation-based inversion without requiring large-scale offline pretraining datasets, Gaussian initialization, and predefined noise schedules.
SFM-FWI trains a single flow network \emph{online} using gradients from the FWI data-misfit, enabling the current inversion model to be treated directly as the starting point of transport dynamics and yielding a stable coarse-to-fine reconstruction trajectory.
Across challenging synthetic benchmarks (Marmousi, Overthrust, and a salt model) and additional stress tests with poor initial models, noisy observations, and sparse acquisition, SFM-FWI consistently improves reconstruction accuracy, structural fidelity, and convergence stability compared with conventional FWI, TV-regularized FWI, and DIP-assisted baselines, while incurring only marginal computational overhead.
These results suggest that physics-driven online flow learning is a practical approach for robust, high-resolution seismic inversion and can be combined with pretrained generative models when such priors are available to provide additional geological knowledge.

\section*{Acknowledgments}
This work was supported by the US Department of Energy under the Advanced Scientific Computing
Research program (grant DE-SC0024563), and the Penn AI Fellowship program.

\bibliographystyle{abbrvnat}
\bibliography{references}

\begin{thebibliography}{45}
\providecommand{\natexlab}[1]{#1}
\providecommand{\url}[1]{\texttt{#1}}
\expandafter\ifx\csname urlstyle\endcsname\relax
  \providecommand{\doi}[1]{doi: #1}\else
  \providecommand{\doi}{doi: \begingroup \urlstyle{rm}\Url}\fi

\bibitem[Alkhalifah et~al.(2018)Alkhalifah, Sun, and Wu]{alkhalifah2018full}
T.~Alkhalifah, B.~B. Sun, and Z.~Wu.
\newblock Full model wavenumber inversion: Identifying sources of information for the elusive middle model wavenumbers.
\newblock \emph{Geophysics}, 83\penalty0 (6):\penalty0 R597--R610, 2018.

\bibitem[Aminzadeh et~al.(1996)Aminzadeh, Burkhard, Long, Kunz, and Duclos]{aminzadeh1996three}
F.~Aminzadeh, N.~Burkhard, J.~Long, T.~Kunz, and P.~Duclos.
\newblock Three dimensional seg/eaeg models—an update.
\newblock \emph{The Leading Edge}, 15\penalty0 (2):\penalty0 131--134, 1996.

\bibitem[Aster et~al.(2011)Aster, Borchers, and Thurber]{Aster2011}
R.~Aster, B.~Borchers, and C.~Thurber.
\newblock \emph{Tikhonov regularization.Parameter estimation and inverse problems. 2nd edn}.
\newblock Elsevier Academic Press, New York, 2011.

\bibitem[Billette and Brandsberg-Dahl(2005)]{billette20052004}
F.~Billette and S.~Brandsberg-Dahl.
\newblock The 2004 bp velocity benchmark.
\newblock In \emph{67th EAGE Conference \& Exhibition}, pages cp--1. European Association of Geoscientists \& Engineers, 2005.

\bibitem[Brossier et~al.(2010)Brossier, Operto, and Virieux]{Brossier2010}
R.~Brossier, S.~Operto, and J.~Virieux.
\newblock Which data residual norm for robust elastic frequency-domain full waveform inversion?
\newblock \emph{GEOPHYSICS}, 75\penalty0 (3):\penalty0 R37--R46, May 2010.
\newblock ISSN 0016-8033.
\newblock \doi{10.1190/1.3379323}.
\newblock URL \url{http://library.seg.org/doi/10.1190/1.3379323}.

\bibitem[Bunks et~al.(1995)Bunks, Saleck, Zaleski, and Chavent]{Bunks1995}
C.~Bunks, F.~M. Saleck, S.~Zaleski, and G.~Chavent.
\newblock Multiscale seismic waveform inversion.
\newblock \emph{GEOPHYSICS}, 60\penalty0 (5):\penalty0 1457--1473, Sept. 1995.
\newblock ISSN 0016-8033.
\newblock \doi{10.1190/1.1443880}.
\newblock URL \url{http://library.seg.org/doi/10.1190/1.1443880}.

\bibitem[Dessa et~al.(2004)Dessa, Operto, Kodaira, Nakanishi, Pascal, Uhira, and Kaneda]{Dessa2004}
J.-X. Dessa, S.~Operto, S.~Kodaira, A.~Nakanishi, G.~Pascal, K.~Uhira, and Y.~Kaneda.
\newblock Multiscale seismic imaging of the eastern {Nankai} trough by full waveform inversion.
\newblock \emph{Geophysical Research Letters}, 31\penalty0 (18):\penalty0 L18606, 2004.
\newblock ISSN 0094-8276.
\newblock \doi{10.1029/2004GL020453}.
\newblock URL \url{http://doi.wiley.com/10.1029/2004GL020453}.

\bibitem[Dhariwal and Nichol(2021)]{dhariwal_diffusion_2021}
P.~Dhariwal and A.~Nichol.
\newblock Diffusion {Models} {Beat} {GANs} on {Image} {Synthesis}, June 2021.
\newblock URL \url{http://arxiv.org/abs/2105.05233}.
\newblock arXiv:2105.05233 [cs, stat].

\bibitem[Fichtner(2010)]{fichtner2010full}
A.~Fichtner.
\newblock \emph{Full seismic waveform modelling and inversion}.
\newblock Springer Science \& Business Media, 2010.

\bibitem[Fichtner et~al.(2013)Fichtner, Trampert, Cupillard, Saygin, Taymaz, Capdeville, and Villasenor]{fichtner2013multiscale}
A.~Fichtner, J.~Trampert, P.~Cupillard, E.~Saygin, T.~Taymaz, Y.~Capdeville, and A.~Villasenor.
\newblock Multiscale full waveform inversion.
\newblock \emph{Geophysical Journal International}, 194\penalty0 (1):\penalty0 534--556, 2013.

\bibitem[Ho et~al.(2020)Ho, Jain, and Abbeel]{ho_denoising_2020}
J.~Ho, A.~Jain, and P.~Abbeel.
\newblock Denoising {Diffusion} {Probabilistic} {Models}.
\newblock In \emph{34th {Conference} on {Neural} {Information} {Processing} {Systems}}, page~25, 2020.

\bibitem[Huang et~al.(2019)Huang, Liu, and Wang]{huang2019robust}
X.~Huang, Y.~Liu, and F.~Wang.
\newblock A robust full waveform inversion using dictionary learning.
\newblock In \emph{SEG International Exposition and Annual Meeting}, page 5407. OnePetro, 2019.

\bibitem[Komatitsch and Martin(2007)]{komatitsch2007unsplit}
D.~Komatitsch and R.~Martin.
\newblock An unsplit convolutional perfectly matched layer improved at grazing incidence for the seismic wave equation.
\newblock \emph{Geophysics}, 72\penalty0 (5):\penalty0 SM155--SM167, 2007.

\bibitem[Lecun et~al.(2015)Lecun, Bengio, and Hinton]{Lecun2015}
Y.~Lecun, Y.~Bengio, and G.~Hinton.
\newblock Deep learning.
\newblock \emph{Nature}, 521\penalty0 (7553):\penalty0 436--444, 2015.
\newblock ISSN 14764687.
\newblock \doi{10.1038/nature14539}.

\bibitem[Levander(1988)]{levander1988fourth}
A.~R. Levander.
\newblock Fourth-order finite-difference p-sv seismograms.
\newblock \emph{Geophysics}, 53\penalty0 (11):\penalty0 1425--1436, 1988.

\bibitem[Li et~al.(2012)Li, Aravkin, {Van Leeuwen}, and Herrmann]{Lix2012}
X.~Li, A.~Y. Aravkin, T.~{Van Leeuwen}, and F.~J. Herrmann.
\newblock {Fast randomized full-waveform inversion with compressive sensing}.
\newblock \emph{Geophysics}, 77\penalty0 (3), 2012.
\newblock ISSN 00168033.
\newblock \doi{10.1190/geo2011-0410.1}.

\bibitem[Lipman et~al.(2022)Lipman, Chen, Ben-Hamu, Nickel, and Le]{lipman2022flow}
Y.~Lipman, R.~T. Chen, H.~Ben-Hamu, M.~Nickel, and M.~Le.
\newblock Flow matching for generative modeling.
\newblock \emph{arXiv preprint arXiv:2210.02747}, 2022.

\bibitem[Loshchilov and Hutter(2017)]{loshchilov2017decoupled}
I.~Loshchilov and F.~Hutter.
\newblock Decoupled weight decay regularization.
\newblock \emph{arXiv preprint arXiv:1711.05101}, 2017.

\bibitem[Luo et~al.(2025)Luo, Huang, and Yang]{luo_self-diffusion_2025}
G.~Luo, S.~Huang, and Y.~Yang.
\newblock Self-diffusion for {Solving} {Inverse} {Problems}, Oct. 2025.
\newblock URL \url{http://arxiv.org/abs/2510.21417}.
\newblock arXiv:2510.21417 [cs].

\bibitem[Mosser et~al.(2020)Mosser, Dubrule, and Blunt]{mosser2020stochastic}
L.~Mosser, O.~Dubrule, and M.~J. Blunt.
\newblock Stochastic seismic waveform inversion using generative adversarial networks as a geological prior.
\newblock \emph{Mathematical Geosciences}, 52\penalty0 (1):\penalty0 53--79, 2020.

\bibitem[Métivier et~al.(2016)Métivier, Brossier, Mérigot, Oudet, and Virieux]{Metivier2016}
L.~Métivier, R.~Brossier, Q.~Mérigot, E.~Oudet, and J.~Virieux.
\newblock Measuring the misfit between seismograms using an optimal transport distance: application to full waveform inversion.
\newblock \emph{Geophysical Journal International}, 205\penalty0 (1):\penalty0 345--377, Apr. 2016.
\newblock ISSN 0956-540X.
\newblock \doi{10.1093/gji/ggw014}.
\newblock URL \url{https://academic.oup.com/gji/article-lookup/doi/10.1093/gji/ggw014}.

\bibitem[Plessix(2006)]{Plessix2006}
R.-E. Plessix.
\newblock A review of the adjoint-state method for computing the gradient of a functional with geophysical applications.
\newblock \emph{Geophysical Journal International}, 167\penalty0 (2):\penalty0 495--503, Nov. 2006.
\newblock ISSN 0956540X.
\newblock \doi{10.1111/j.1365-246X.2006.02978.x}.
\newblock URL \url{https://academic.oup.com/gji/article-lookup/doi/10.1111/j.1365-246X.2006.02978.x}.

\bibitem[Pratt(1990)]{Pratt1990}
R.~G. Pratt.
\newblock Inverse theory applied to multi-source cross-hole tomography. {Part} 2: elastic wave-equation method.
\newblock \emph{Geophysical Prospecting}, 38\penalty0 (3):\penalty0 311--329, 1990.

\bibitem[Richardson(2026)]{richardson_alan_2026}
A.~Richardson.
\newblock Deepwave, Jan. 2026.
\newblock URL \url{https://doi.org/10.5281/zenodo.3829886}.

\bibitem[Sirgue and Pratt(2004)]{sirgue2004efficient}
L.~Sirgue and R.~G. Pratt.
\newblock Efficient waveform inversion and imaging: A strategy for selecting temporal frequencies.
\newblock \emph{Geophysics}, 69\penalty0 (1):\penalty0 231--248, 2004.

\bibitem[Song et~al.(2023)Song, Wang, Richardson, and Liu]{song2023weighted}
C.~Song, Y.~Wang, A.~Richardson, and C.~Liu.
\newblock Weighted envelope correlation-based waveform inversion using automatic differentiation.
\newblock \emph{IEEE Transactions on Geoscience and Remote Sensing}, 61:\penalty0 1--11, 2023.

\bibitem[Song et~al.(2021)Song, Sohl-Dickstein, Kingma, Kumar, Ermon, and Poole]{song_score-based_2021}
Y.~Song, J.~Sohl-Dickstein, D.~P. Kingma, A.~Kumar, S.~Ermon, and B.~Poole.
\newblock Score-{Based} {Generative} {Modeling} through {Stochastic} {Differential} {Equations}, Feb. 2021.
\newblock URL \url{http://arxiv.org/abs/2011.13456}.
\newblock arXiv:2011.13456 [cs, stat].

\bibitem[Sun et~al.(2023{\natexlab{a}})Sun, Innanen, Zhang, and Trad]{sun2023implicit}
J.~Sun, K.~Innanen, T.~Zhang, and D.~Trad.
\newblock Implicit seismic full waveform inversion with deep neural representation.
\newblock \emph{Journal of Geophysical Research: Solid Earth}, 128\penalty0 (3):\penalty0 e2022JB025964, 2023{\natexlab{a}}.

\bibitem[Sun et~al.(2023{\natexlab{b}})Sun, Yang, Liang, and Ma]{sun2023full}
P.~Sun, F.~Yang, H.~Liang, and J.~Ma.
\newblock Full-waveform inversion using a learned regularization.
\newblock \emph{IEEE Transactions on Geoscience and Remote Sensing}, 61:\penalty0 1--15, 2023{\natexlab{b}}.

\bibitem[Symes(2008)]{symes2008migration}
W.~W. Symes.
\newblock Migration velocity analysis and waveform inversion.
\newblock \emph{Geophysical prospecting}, 56\penalty0 (6):\penalty0 765--790, 2008.

\bibitem[Tarantola(1984)]{Tarantola1984}
A.~Tarantola.
\newblock Inversion of seismic reflection data in the acoustic approximation.
\newblock \emph{GEOPHYSICS}, 49\penalty0 (8):\penalty0 1259--1266, Aug. 1984.
\newblock ISSN 0016-8033.
\newblock \doi{10.1190/1.1441754}.
\newblock URL \url{http://library.seg.org/doi/10.1190/1.1441754}.

\bibitem[Tromp(2020)]{tromp2020seismic}
J.~Tromp.
\newblock Seismic wavefield imaging of earth’s interior across scales.
\newblock \emph{Nature Reviews Earth \& Environment}, 1\penalty0 (1):\penalty0 40--53, 2020.

\bibitem[Ulyanov et~al.(2018)Ulyanov, Vedaldi, and Lempitsky]{ulyanov2018deep}
D.~Ulyanov, A.~Vedaldi, and V.~Lempitsky.
\newblock Deep image prior.
\newblock In \emph{Proceedings of the IEEE conference on computer vision and pattern recognition}, pages 9446--9454, 2018.

\bibitem[Venkatakrishnan et~al.(2013)Venkatakrishnan, Bouman, and Wohlberg]{venkatakrishnan2013plug}
S.~V. Venkatakrishnan, C.~A. Bouman, and B.~Wohlberg.
\newblock Plug-and-play priors for model based reconstruction.
\newblock In \emph{2013 IEEE global conference on signal and information processing}, pages 945--948. IEEE, 2013.

\bibitem[Versteeg(1994)]{versteeg1994marmousi}
R.~Versteeg.
\newblock The marmousi experience: Velocity model determination on a synthetic complex data set.
\newblock \emph{The Leading Edge}, 13\penalty0 (9):\penalty0 927--936, 1994.

\bibitem[Virieux and Operto(2009)]{Virieux2009}
J.~Virieux and S.~Operto.
\newblock An overview of full-waveform inversion in exploration geophysics.
\newblock \emph{GEOPHYSICS}, 74\penalty0 (6):\penalty0 WCC1--WCC26, Nov. 2009.
\newblock ISSN 0016-8033.
\newblock \doi{10.1190/1.3238367}.
\newblock URL \url{https://library.seg.org/doi/10.1190/1.3238367}.

\bibitem[Wang et~al.(2023)Wang, Huang, and Alkhalifah]{wang_prior_2023}
F.~Wang, X.~Huang, and T.~A. Alkhalifah.
\newblock A {Prior} {Regularized} {Full} {Waveform} {Inversion} {Using} {Generative} {Diffusion} {Models}.
\newblock \emph{IEEE Transactions on Geoscience and Remote Sensing}, 61:\penalty0 1--11, 2023.
\newblock ISSN 1558-0644.
\newblock \doi{10.1109/TGRS.2023.3337014}.
\newblock URL \url{https://ieeexplore.ieee.org/document/10328845}.
\newblock Conference Name: IEEE Transactions on Geoscience and Remote Sensing.

\bibitem[Wang et~al.(2025)Wang, Huang, and Alkhalifah]{wang_geological_2025}
F.~Wang, X.~Huang, and T.~Alkhalifah.
\newblock Geological and {Well} prior assisted full waveform inversion using conditional diffusion models, July 2025.
\newblock URL \url{http://arxiv.org/abs/2412.06959}.
\newblock arXiv:2412.06959 [physics].

\bibitem[Wang et~al.(2004)Wang, Bovik, Sheikh, and Simoncelli]{wang2004image}
Z.~Wang, A.~C. Bovik, H.~R. Sheikh, and E.~P. Simoncelli.
\newblock Image quality assessment: from error visibility to structural similarity.
\newblock \emph{IEEE transactions on image processing}, 13\penalty0 (4):\penalty0 600--612, 2004.

\bibitem[Warner et~al.(2013)Warner, Ratcliffe, Nangoo, Morgan, Umpleby, Shah, Vinje, Štekl, Guasch, Win, Conroy, and Bertrand]{Warner2013}
M.~Warner, A.~Ratcliffe, T.~Nangoo, J.~Morgan, A.~Umpleby, N.~Shah, V.~Vinje, I.~Štekl, L.~Guasch, C.~Win, G.~Conroy, and A.~Bertrand.
\newblock Anisotropic {3D} full-waveform inversion.
\newblock \emph{GEOPHYSICS}, 78\penalty0 (2):\penalty0 R59--R80, Mar. 2013.
\newblock ISSN 0016-8033.
\newblock \doi{10.1190/geo2012-0338.1}.
\newblock URL \url{http://library.seg.org/doi/10.1190/geo2012-0338.1}.

\bibitem[Wu and Ma(2025)]{wu2025does}
Y.~Wu and J.~Ma.
\newblock How does neural network reparametrization improve geophysical inversion?
\newblock \emph{Journal of Geophysical Research: Machine Learning and Computation}, 2\penalty0 (2):\penalty0 e2025JH000621, 2025.

\bibitem[Wu and McMechan(2019)]{wu2019parametric}
Y.~Wu and G.~A. McMechan.
\newblock Parametric convolutional neural network-domain full-waveform inversion.
\newblock \emph{Geophysics}, 84\penalty0 (6):\penalty0 R881--R896, 2019.

\bibitem[Xiang and Zhang(2016)]{xiang2016efficient}
S.~Xiang and H.~Zhang.
\newblock Efficient edge-guided full-waveform inversion by canny edge detection and bilateral filtering algorithms.
\newblock \emph{Geophysical Supplements to the Monthly Notices of the Royal Astronomical Society}, 207\penalty0 (2):\penalty0 1049--1061, 2016.

\bibitem[Xue et~al.(2017)Xue, Zhu, and Fomel]{Xue2017}
Z.~Xue, H.~Zhu, and S.~Fomel.
\newblock Full-waveform inversion using seislet regularization.
\newblock \emph{Geophysics}, 82\penalty0 (5):\penalty0 A43--A49, 2017.
\newblock ISSN 19422156.
\newblock \doi{10.1190/GEO2016-0699.1}.

\bibitem[Zhu et~al.(2022)Zhu, Xu, Darve, Biondi, and Beroza]{zhu2022integrating}
W.~Zhu, K.~Xu, E.~Darve, B.~Biondi, and G.~C. Beroza.
\newblock Integrating deep neural networks with full-waveform inversion: Reparameterization, regularization, and uncertainty quantification.
\newblock \emph{Geophysics}, 87\penalty0 (1):\penalty0 R93--R109, 2022.

\end{thebibliography}

\end{document}